\documentclass[10pt,twocolumn,letterpaper]{article}

\usepackage{cvpr}              

\usepackage{graphicx}

\usepackage{amsmath,amsfonts,bm,amsthm}

\newtheorem*{remark}{Remark}

\def\eqref#1{equation~\ref{#1}}

\def\1{\bm{1}}

\DeclareMathAlphabet{\mathsfit}{\encodingdefault}{\sfdefault}{m}{sl}
\SetMathAlphabet{\mathsfit}{bold}{\encodingdefault}{\sfdefault}{bx}{n}

\usepackage{algorithm}
\usepackage{algpseudocode}
\usepackage{wrapfig}
\usepackage{url}
\usepackage[acronym]{glossaries}

\newacronym{clip}{CLIP}{Contrastive Language-Image Pretraining}
\newacronym{svd}{SVD}{Singular Value Decomposition}
\newacronym{cnn}{CNN}{Convolutional Neural Net}
\newacronym{coop}{CoOp}{Context Optimization}
\newacronym{mle}{MLE}{Maximum Likelihood Estimation}
\newacronym{map}{MAP}{Maximum A Posteriori}
\newacronym{sde}{SDE}{Stochastic Differential Equation}
\newacronym{sgld}{SGLD}{Stochastic Gradient Langevin Dynamics}
\newacronym{sghmc}{SGHMC}{Stochastic Gradient Hamiltonian Monte Carlo}
\newacronym{csghmc}{cSGHMC}{Cyclical SGHMC}
\newacronym{cocoop}{CoCoOp}{Conditional CoOp}
\newacronym{maple}{MaPLe}{Multi-modal Prompt Learning}
\newacronym{sgmcmc}{SGMCMC}{Stochastic Gradient Markov Chain Monte Carlo}
\newacronym{rcsgmcmc}{rcSGMCMC}{Repulsive Cyclical SGMCMC}
\newacronym{mmd}{MMD}{Maximum Mean Discrepancy}
\newacronym{ot}{OT}{Optimal Transport}

\usepackage{booktabs}        
\usepackage{xcolor}          
\usepackage[table]{xcolor}   
\usepackage{subcaption}      
\usepackage{colortbl}        

\definecolor{MidnightBlue}{RGB}{25,25,112}  
\definecolor{Bittersweet}{RGB}{254,97,0}    
\definecolor{tabhighlight}{RGB}{240,240,240} 

\usepackage{sidecap}

\usepackage{float}

\usepackage{fontawesome}

\definecolor{cvprblue}{rgb}{0.21,0.49,0.74}
\definecolor{cvprred}{rgb}{0.74, 0.22, 0.22}
\definecolor{cvprgreen}{rgb}{0.00, 0.47, 0.32}
\usepackage[pagebackref,breaklinks,colorlinks,allcolors=cvprblue]{hyperref}

\def\paperID{redacted} 
\def\confName{CVPR}
\def\confYear{2026}

\title{ReBaPL: Repulsive Bayesian Prompt Learning}

\author{
Yassir Bendou$^{1,\star}$, Omar Ezzahir$^{1,\star}$, Eduardo Montesuma$^{1,\star,\dagger}$\\ Gabriel Mahuas$^{1}$, Victoria Schevchenko$^{1}$, Mike Gartrell$^{2,\dagger}$\\
$^{1}$ Sigma Nova, Paris, France\quad$^{2}$Rhizome Labs, Paris, France\\
\texttt{first-name.second-name@sigmanova.ai}\\ \texttt{first-name.second-name@rhizome-labs.com}
}

\begin{document}
\maketitle
\begingroup
\renewcommand\thefootnote{}

\footnotetext{$^{\star}$ Equal contribution.}
\footnotetext{$^{\dagger}$ Corresponding author.}
\footnotetext{Our code is available at \faGithub\,\url{https://github.com/SigmaNova/ReBaPL}}

\endgroup
\begin{abstract}
Prompt learning has emerged as an effective technique for fine-tuning large-scale foundation models for downstream tasks. However, conventional prompt learning methods are prone to overfitting and can struggle with out-of-distribution generalization. To address these limitations, Bayesian prompt learning has been proposed, which frames prompt optimization as a Bayesian inference problem to enhance robustness. This paper introduces Repulsive Bayesian Prompt Learning (ReBaPL), a novel method for Bayesian prompt learning, designed to efficiently explore the complex and often multimodal posterior landscape of prompts. Our method integrates a cyclical step-size schedule with a stochastic gradient Hamiltonian Monte Carlo (SGHMC) algorithm, enabling alternating phases of exploration to discover new modes, and exploitation to refine existing modes. Furthermore, we introduce a repulsive force derived from a potential function over probability metrics (including Maximum Mean Discrepancy and Wasserstein distance) computed on the distributions of representations produced by different prompts. This representation-space repulsion diversifies exploration and prevents premature collapse to a single mode. Our approach allows for a more comprehensive characterization of the prompt posterior distribution, leading to improved generalization. In contrast to prior Bayesian prompt learning methods, our method provides a modular plug-and-play Bayesian extension of any existing prompt learning method based on maximum likelihood estimation. We demonstrate the efficacy of ReBaPL on several benchmark datasets, showing superior performance over state-of-the-art prompt learning methods.
\end{abstract}
    
\section{Introduction}

Large-scale vision-language models (VLMs), such as CLIP~\citep{radford2021learning}, have demonstrated remarkable zero-shot generalization capabilities across a wide array of visual tasks. Many works have shown better performance on downstream task by adapting VLMs for few-shot image classification~\cite{zhou2022learning, zhou2022conditional, clipadapter, tipadapter, proker}. Among these methods, prompt learning has emerged as an efficient model adaptation technique, enabling VLMs and other foundation models to be tailored for specific downstream tasks by learning continuous prompt vectors, instead of fine-tuning the entire model. However, this approach is not without its drawbacks. Standard prompt learning, which typically optimizes prompts through maximum likelihood estimation (MLE), is prone to overfitting on training data, leading to diminished generalization performance on out-of-distribution (OOD) samples and unseen classes.

The initial MLE-based prompt learning method, known as CoOp~\citep{zhou2022learning}, is especially prone to overfitting \cite{Ma2022-kw}. In an attempt to mitigate this issue, a subsequent MLE-based prompt learning method, known as CoCoOp~\citep{zhou2022conditional}, proposed learning an instance-specific continuous prompt that is conditioned on the input image. While CoCoOp performs better than CoOp, it can still suffer from generalization issues. 

In contrast to prompt learning for the text modality only, other recent approaches involve multi-modal prompt learning for both image and text representations~\citep{khattak2023maple, guo2025mmrl, chengvamp2025, Yang2025-fi}. These approaches introduce learnable prompt tokens at varying depths in the transformer layers in the image and text encoders, and then generally use a vision-language coupling function to induce learning of prompts in a shared embedding space. 

Other MLE prompt learning methods~\citep{lu2022prompt, khattak2023self} generally focus on regularization-based approaches to mitigate overfitting \cite{zhou2022conditional, Zhu2025-jq}. One such approach, PromptSRC~\citep{khattak2023self}, uses self-regulating constraints to prevent prompts from losing the generalized knowledge of the pretrained model. The ProDA method~\citep{lu2022prompt} is a probabilistic approach, where the prompt distribution is fit to a Gaussian distribution using MLE, with a regularization term to improve prompt diversity.  While effective at mitigating overfitting, these methods guide the learning process towards a single optimal solution manifold and may not fully capture the complex, potentially multi-modal posterior distribution of effective prompts. 

Rather than seeking a single point estimate with regularization, an alternative paradigm is to characterize the full distribution over prompts through Bayesian inference. Bayesian prompt learning methods~\citep{derakhshani2023bayesian, cho2024make, kim2025bayesian, Qu2025-rx, chengvamp2025} 
frame prompt learning as a Bayesian inference problem. These methods aim to enhance robustness and improve generalization by inferring a posterior distribution over the prompt space. This probabilistic perspective naturally introduces regularization, which helps prevent the model from learning spurious features and overfitting to the training set. Most Bayesian prompt learning approaches estimate the posterior using a variational unimodal Gaussian approximation, which limits diversity. To address this limitation, the Adaptive Particle-based Prompt Learning (APP) approach~\citep{cho2024make} uses a Wasserstein gradient flow and Stein Variational Gradient Descent (SVGD) to approximate multiple modes in the posterior using a variational distribution represented by a collection of interacting particles. VaMP~\cite{chengvamp2025} extends multi-modal (text-image) prompt learning by introducing variational inference to model prompts as probabilistic latent variables rather than deterministic parameters. This enables instance-specific, uncertainty-aware prompt generation, where text prompts are dynamically conditioned on input image features and sampled from learned posterior distributions, with a class-aware prior providing semantic regularization. While VaMP incorporates uncertainty modeling into multi-modal prompting, it relies on a variational approximation with a unimodal Gaussian posterior, which may limit its ability to capture the full complexity of multimodal prompt distributions.

We propose a novel Repulsive Bayesian Prompt Learning (ReBaPL) approach based on cyclical stochastic gradient Hamiltonian Monte Carlo (rcSGHMC).
In contrast to the deterministic SVGD method used in APP, our approach leverages an efficient MCMC algorithm where the posterior is represented as a collection of samples, coupled to a repulsive force based on the interaction between prompt representations.
This allows our method to provide a richer representation of the shape of the high-density regions around multiple modes in the posterior, which results in better generalization to novel classes without overfitting on the base classes. In contrast to prior Bayesian prompt learning methods, our method provides a modular plug-and-play Bayesian extension of any existing MLE-based prompt learning method. We implement and perform experiments with ReBaPL running as a plug-and-play learning algorithm on top of two existing prompt learning methods: MaPLe~\citep{khattak2023maple} and MMRL~\citep{guo2025mmrl}.

Our contributions include:

\begin{itemize}
    \item Repulsive cyclical SGHMC: We propose the rcSGHMC algorithm for approximating complex multimodal posteriors. Through the use of Hamiltonian dynamics, cyclical learning rates, and repulsion between prompt representations, our method is effective at exploring the complex posteriors with multiple modes that are found in prompt learning.
    \item Representation-based repulsion: Rather than comparing parameters directly in weight space, our method introduces a repulsive potential based on probability metrics, including Maximum Mean Discrepancy (MMD) and Wasserstein distance, between the distributions of representations. This allows us to capture the functional similarity between prompts through their induced representations, encouraging exploration of functionally diverse modes in the posterior.
    \item We show experimentally that our approach provides rich characterization of the space of diverse prompts, which improves generalization performance on base-to-novel tasks, cross-dataset transfer, and domain generalization.
\end{itemize}

\section{Background}

\begin{figure*}[ht]
    \centering
    \includegraphics[width=0.9\textwidth]{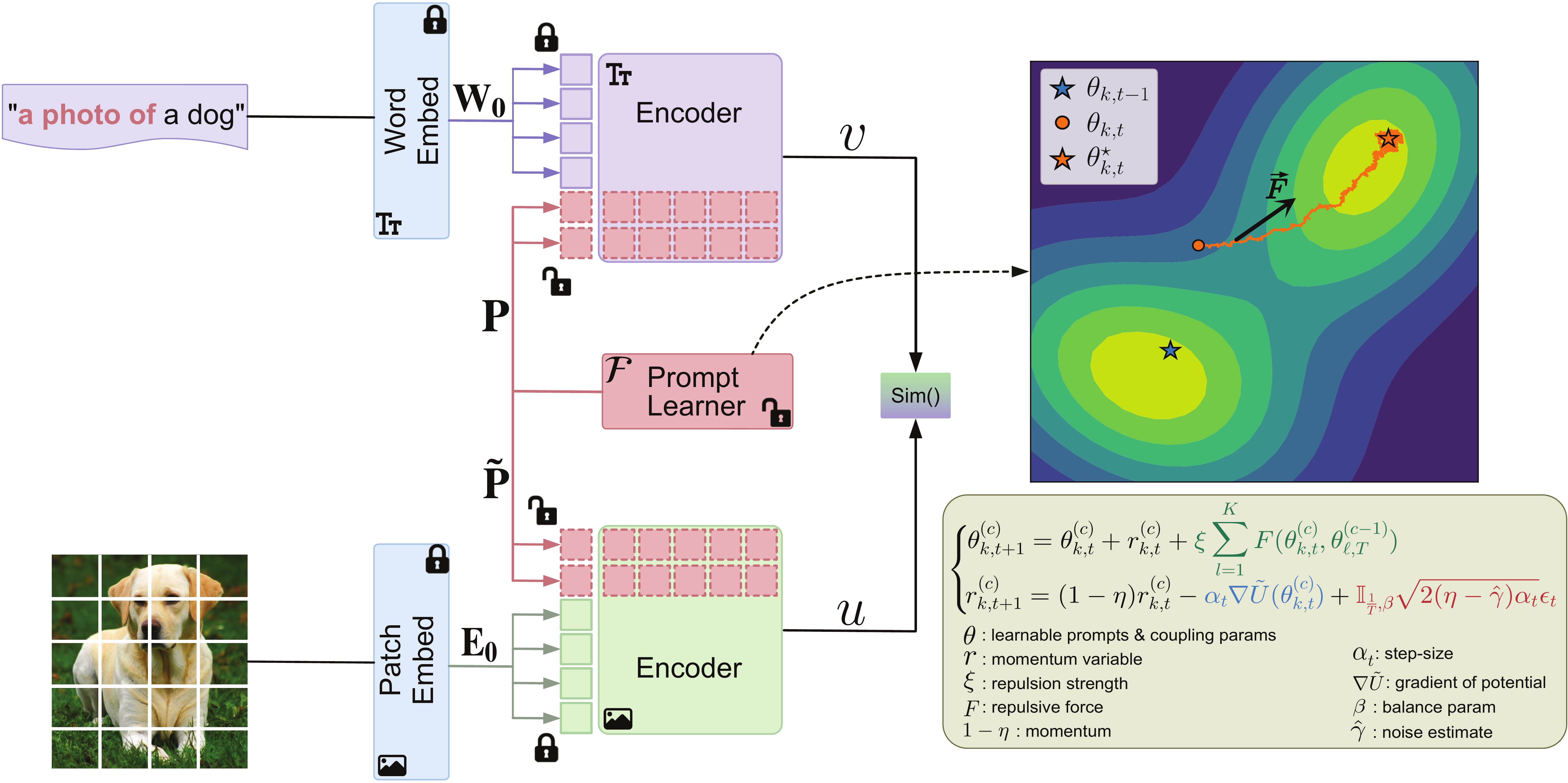}
    \caption{Overview of our proposed ReBaPL approach, in a multi-modal prompt learning setting. Text and image encoders receive text ($\mathbf{W_0}$) and image ($\mathbf{E_0}$) embeddings as input, combined with learnable tokens of context prompts ($\mathbf{P}$). The terms pertaining to the exploration and sampling stages and the repulsion force are colored in \textcolor{cvprblue}{blue}, \textcolor{cvprred}{red} and \textcolor{cvprgreen}{green} respectively.}
    \label{fig:illustration}
\end{figure*}

In this section we introduce the core principles of our ReBaPL method, namely: prompt learning (Section~\ref{sec:prompt-learning}), Bayesian learning (Section~\ref{sec:bayesian-learning}), and probability metrics (Section~\ref{sec:prob-metrics}). Figure~\ref{fig:illustration} provides an overview of our approach.

\subsection{Prompt Learning}\label{sec:prompt-learning}

\gls{clip}~\citep{radford2021learning} is a pre-training method that learns a joint latent space for texts and images. Its main principle is \emph{encoding} an image through a \gls{cnn}, e.g., a ResNet~\citep{he2016deep}) or a Transformer, e.g., a ViT~\citep{dosovitskiy2020image}, and a text snippet through a Transformer~\citep{vaswani2017attention}. From a pre-trained \gls{clip}, it is possible to perform downstream tasks, such as classification. For example, given an image $x$, one can measure the alignment with a textual prompt $T$ in the latent space. A good candidate for the textual prompt is \emph{"A photo of [CLS]"}, where \emph{"[CLS]"} corresponds to the class with which one is measuring the alignment. In this sense, \gls{clip} already demonstrates its own \emph{zero-shot} generalization ability, mainly due to pre-training. Prompt learning takes this idea further, where the prompt is treated as an \emph{optimization variable during learning}~\citep{zhou2022learning}.

Early prompt learning methods for VLMs, such as CoOp~\citep{zhou2022learning} and CoCoOp~\citep{zhou2022conditional}, focus exclusively on learning prompts in the \emph{language branch} of \gls{clip}. While these approaches demonstrate improved performance on downstream tasks, they adopt a \emph{unimodal} prompting strategy that only partially adapts the pre-trained model. This limitation motivated the development of \emph{multi-modal prompt learning}, which recognizes that both the vision and language encoders should be adapted simultaneously to achieve optimal alignment between modalities~\citep{khattak2023maple}. While our ReBaPL approach can be run on top of any MLE-based prompt learning method, we focus primarily on multi-modal prompt learning for the remainder of this paper, since these methods tend to significantly outperform unimodal methods in terms of predictive performance~\cite{khattak2023maple, guo2025mmrl}. 

\textbf{Multi-modal Prompt Learning} Given \gls{clip}'s dual-encoder architecture with text encoder $\mathcal{L}$ and image encoder $\mathcal{V}$, multi-modal prompt learning approaches~\citep{khattak2023maple, guo2025mmrl, chengvamp2025} introduce learnable context tokens in \emph{both} branches at multiple depths. Let $\mathcal{V} = \{V_{i}\}_{i=1}^{K}$ and $\mathcal{L} = \{L_{i}\}_{i=1}^{K}$ denote the $K$ transformer layers in the vision and language branches, respectively. In the case of MaPLe\citep{khattak2023maple}, MMRL~\citep{guo2025mmrl}, and VaMP~\citep{chengvamp2025}, prompt tuning is further implemented in deeper layers in the encoders.

For the \emph{language branch}, learnable tokens $\{P_{i} \in \mathbb{R}^{d_{\text{text}}}\}_{i=1}^{b}$ are introduced alongside the input word embeddings $W_0 = [w_0^1, w_0^2, \ldots, w_0^N] \in \mathbb{R}^{N \times d_{\text{text}}}$, forming the input $[P^1, P^2, \ldots, P^b, W_0]$ to the first transformer layer. Additional learnable tokens are introduced in the transformer blocks of the language encoder $\mathcal{L}_i$, in deeper layers up to depth $J < K$:
\begin{align}
    [\_, W_i] &= \mathcal{L}_i([P_{i-1}, W_{i-1}]) \quad i = 1, 2, \ldots, J,
\end{align}
where the learned prompts $P_i$ are processed at each layer, and $[\cdot, \cdot]$ is the concatenation operation. After depth $J$, subsequent layers process the prompts from layer $J$, and the final text representation $v$ is computed by projecting the text embeddings to a common vision-language embedding space:
\begin{align}
    [P_j, W_j] &= \mathcal{L}_j([P_{j-1}, W_{j-1}]) \quad j = J+1, \ldots, K, \\
    v &= \text{TextProj}(w^N_K) 
\end{align}

Similarly, for the \emph{vision branch}, learnable tokens $\{\tilde{P}_{i} \in \mathbb{R}^{d_{\text{vis}}}\}_{i=1}^{b}$ are introduced alongside the patch embeddings $E_0$ and class token $c_0$. The vision prompts are processed through transformer layers of the vision encoder $\mathcal{V}_i$ analogously to the language prompts, with learnable tokens introduced up to depth $J$, where $u$ is the final image representation is obtained by projecting to a common vision-language embedding space:
\begin{align}
    [c_i, E_i, \_] &= \mathcal{V}_i([c_{i-1}, E_{i-1}, \tilde{P}_{i-1}]) \quad i = 1, 2, \ldots, J, \\
    [c_j, E_j, \tilde{P}_j] &= \mathcal{V}_j([c_{j-1}, E_{j-1}, \tilde{P}_{j-1}]) \quad j = J + 1, \ldots, K, \\
    u &= \text{ImageProj}(c_K)
\end{align}

\textbf{Vision-Language Coupling.} A key insight of recent multi-modal prompt learning is that the vision and language prompts should not be learned independently~\cite{khattak2023maple, guo2025mmrl}. To ensure mutual synergy between modalities, a \emph{coupling function} $\mathcal{F}$ explicitly conditions the vision prompts on their language counterparts:
\begin{align}
    \tilde{P}_i = \mathcal{F}_i(P_i),
\end{align}
where the implementation of $\mathcal{F}_i: \mathbb{R}^{d_{\text{text}}} \rightarrow \mathbb{R}^{d_{\text{vis}}}$ depends on the particular multi-prompt learning method. For MaPLe~\citep{khattak2023maple}, $\mathcal{F}$ is implemented as a learnable linear projection. In MMRL~\citep{guo2025mmrl}, the coupling function uses visual tokens embedded in a shared latent space, which are initialized by sampling from a Gaussian, and then uses a separate linear projection to generate modality-specific prompts. VaMP~\citep{chengvamp2025} employs a different coupling strategy, where text prompts are generated by layer-specific MLPs that map frozen CLIP image features to prompt tokens, creating sample-specific text prompts, while vision prompts remain shared and learnable across samples. 
For all of these approaches the coupling function acts as a bridge between the two modalities, allowing mutual gradient propagation and promoting synergistic adaptation. This explicit conditioning discourages independent unimodal solutions and encourages learning of prompts in a shared embedding.

Given a dataset $\mathcal{D} = \{x_{i}, y_{i}\}_{i=1}^{n}$ of images $x_{i}$ and labels $y_{i} \in \{1,\cdots,C\}$, multi-modal prompt learning optimizes the language prompts $\{P_i\}$, vision prompts $\{\tilde{P}_i\}$, and coupling functions $\{\mathcal{F}_i\}$ via:
\begin{align}
    \theta^{\star} &= \underset{\theta}{\text{arg min}} -\dfrac{1}{n}\sum_{i=1}^{n}\log p(y_{i}|u_i,\theta), \label{eq:maple}\\
    p(y_{i}|u_i,\theta) &= \dfrac{\exp(\text{cossim}(u_{i}, v_{y_{i}}) / \tau)}{\sum_{c=1}^{C}\exp(\text{cossim}(u_{i}, v_{c}) / \tau)}, \label{eq:predictive-distribution}
\end{align}
where $\theta$ encompasses all learnable prompts and coupling parameters, $u_{i}$ is the image embedding from the vision encoder (now influenced by vision prompts $\tilde{P}_i$), $v_c$ is the text embedding for class $c$ (influenced by language prompts $P_i$), and $\tau$ is a temperature parameter. 


\subsection{Bayesian Learning}\label{sec:bayesian-learning}



Bayesian learning is a framework for reasoning under uncertainty, which is particularly relevant to data-scarce applications~\cite{papamarkou2024position}. Starting from Bayes theorem~\cite{smith1997statistics}, for a dataset $\mathcal{D}$ and a model parameterized by $\theta$, the posterior probability over parameters $p(\theta|\mathcal{D})$ can be modeled as,
\begin{align}
    p(\theta|\mathcal{D}) = \dfrac{p(\mathcal{D}|\theta)p(\theta)}{p(\mathcal{D})},\text{ then } p(\theta|\mathcal{D}) \propto p(\mathcal{D}|\theta)p(\theta).
\end{align}
In practice, we want to estimate $\log p(\theta|\mathcal{D})$, or, equivalently (modulo the normalization constant $p(\mathcal{D})$) $\log p(\mathcal{D}|\theta) + \log p(\theta)$. The first term is the log-likelihood, and the second term is the prior over parameters. Henceforth, we call $U(\theta) = -\log p(\mathcal{D}|\theta)$ the \emph{potential}.

One way to estimate the posterior in Bayesian learning is \emph{sampling} from $p(\theta|\mathcal{D})$, i.e., acquiring samples with high probability under this distribution. This can be done by adding the proper amount of noise to a gradient-based optimization algorithm. Indeed, assuming $\theta_{1},\cdots,\theta_{K} \sim p(\theta)$, we can flow these \emph{samples} with Langevin dynamics,
\begin{align}
    \theta_{k,t+1} = \theta_{k,t} - \alpha_{t}\nabla U(\theta_{k,t}) + \sqrt{2\alpha_{t}}\epsilon_{t},\label{eq:sgld}
\end{align}
where $\epsilon_{t} \sim \mathcal{N}(0, \text{I}_{d})$. This equation corresponds to the discretization in time of the Langevin \gls{sde}, $\dot{\theta}_{k}(t) = -\nabla U(\theta_{k}(t))dt + \sqrt{2}dW_{t}$, where $dW_{t}$ is a standard $d-$dimensional Wiener process. Solving this \gls{sde} is potentially intractable, since computing $\nabla U(\theta_{k}) = - \nabla \log p(\mathcal{D}|\theta)$ involves evaluating the likelihood term over the complete dataset. This motivated~\cite{welling2011bayesian} to propose the \gls{sgld} algorithm, which estimates the potential's gradient, $\nabla U$, over mini-batches from $\mathcal{D}$, that is,
\begin{equation*}
    \tilde{U}(\theta) = -\dfrac{m}{n}\sum_{i=1}^{m}\log p(x_{i}|\theta)+\log p(\theta),
\end{equation*}
where $m \ll n$ is the mini-batch size. However, this algorithm suffers from slow convergence in high dimensions.

To solve the limitations of \gls{sgld},~\cite{sghmc} proposes momentum variables $r$ with a friction term, thus introducing \gls{sghmc},
\begin{equation}
\begin{cases}
    \theta_{k,t+1} = \theta_{k,t} + r_{k,t}\\
    r_{k,t+1}=r_{k,t} - \alpha_{t} \nabla \tilde{U}(\theta_{k,t}) - \eta r_{k,t} + \sqrt{2(\eta - \hat{\gamma})\alpha_{t}}\epsilon_{t}.\label{eq:sghmc}
\end{cases}
\end{equation}
Here, $\hat{\gamma} $ estimates the stochastic gradient noise. The friction term $\eta$ counteracts this noise to ensure convergence to the correct stationary distribution. This allows SGHMC to combine the computational efficiency of mini-batching with the rapid exploration of momentum-based dynamics.

In a further development, ~\cite{zhang2019cyclical} proposed alternating between exploration and sampling stages for enhancing the diversity of samples in \gls{sgmcmc} methods. First, they update the learning rate using a cosine scheduler. Second, they alternate between iterations of exploration and sampling cyclically. This approach is known as Cyclical \gls{sgmcmc}.

\begin{remark}{(Nomenclature of methods)}
    As discussed in~\cite{zhang2019cyclical}, both \gls{sgld} and \gls{sghmc} can be unified under the common framework of \gls{sgmcmc} methods. This nomenclature hints at the fact that these algorithms are stochastic gradient discretizations of continuous-time Markov processes designed to sample from the posterior distribution, differing only in how they incorporate momentum, friction, and injected noise. For the remainder of the paper, we adopt this convention for Bayesian methods.
\end{remark}

\subsection{Probability Metrics}\label{sec:prob-metrics}

In this section, we present metrics between probability distributions that will be used in our method. Recall that, given a subset $\Omega \in \mathbb{R}^{d}$, $\mathcal{P}(\Omega)$ denotes the set of probability distributions on $\Omega$. A probability metric is a metric on elements of $\mathcal{P}(\Omega)$. In the following, we discuss how to estimate these metrics based on finite samples from $p$ and $q$, denoted $\{ z_{i}^{(p)} \}_{i=1}^{n}$ and $\{ z_{j}^{(q)} \}_{j=1}^{m}$, respectively.

The \gls{mmd} metric, proposed by~\cite{gretton2012kernel}, computes a distance between $p$ and $q$ based on a kernel $\kappa:\Omega\times\Omega\rightarrow\mathbb{R}$. More specifically,
\begin{align}
    d(p, q)^{2} = &\dfrac{1}{n^{2}}\sum_{i,j=1}^{n}\kappa(z_{i}^{(p)},z_{j}^{(p)}) + \dfrac{1}{m^{2}}\sum_{i,j=1}^{m}\kappa(z_{i}^{(q)},z_{j}^{(q)})\nonumber\\
    &-\dfrac{2}{nm}\sum_{i=1}^{n}\sum_{j=1}^{m}\kappa(z_{i}^{(p)},z_{j}^{(q)}).\label{eq:mmd}
\end{align}

Meanwhile, the Wasserstein distance comes from \gls{ot}~\cite{montesuma2024recent}. This distance computes the least amount of effort or energy to transport one distribution into another. In mathematical terms,
\begin{align}
    d(p, q)^{2} = \sum_{i=1}^{n}\sum_{j=1}^{m}\gamma_{ij}^{\star}\lVert z_{i}^{(p)} - z_{j}^{(q)} \rVert_{2}^{2},\label{eq:w2}
\end{align}
where $\gamma^{\star} = \text{argmin}_{\gamma \in \Gamma}\sum_{i=1}^{n}\sum_{j=1}^{m}\gamma_{ij}\lVert z_{i}^{(p)} - z_{j}^{(q)} \rVert_{2}^{2}$ is called the \emph{optimal transport plan}. Here, $\Gamma = \{\gamma \in \mathbb{R}^{n\times n}_{+}:\sum_{i=1}^{n} \gamma_{ij} = m^{-1},\sum_{j=1}^{m}\gamma_{ij}=n^{-1}\}$ is the set of feasible transportation plans.

In either of these cases, one can see $d(p, q)$ as a function of its samples, i.e., $\{\{z_{i}^{(p)}\}_{i=1}^{n},\{z_{j}^{(q)}\}_{j=1}^{m}\} \mapsto d(p, q)$. In this sense, Equations~\ref{eq:mmd} and~\ref{eq:w2} compute a distance between groups of points. This will be useful in the next section (c.f. Equation~\ref{eq:dist-params}), as probability metrics will serve as a way of capturing the geometry of the weight space. More details are available in Appendix~\ref{appx:background}.

\section{Bayesian Prompt Learning with Repulsive Cyclical SGHMC}\label{sec:methodology}



Our goal in this section is to present a Bayesian view of prompt learning. We recall that $\mathcal{D} = \{x_{i},y_{i}\}_{i=1}^{n}$ is a dataset of i.i.d. samples. Each text prompt and image in $\mathcal{D}$ is encoded, i.e., $v_{y_{i}} = \text{TextProj}(w^N_K)$ and $u_{i} = \text{ImageProj}(c_K)$. Starting from Equations~\ref{eq:maple} and~\ref{eq:predictive-distribution}, and under the i.i.d. assumption, the log-likelihood is written as $\log p(\mathcal{D}|\theta) = \sum_{i=1}^{n} p(y_{i}|u_{i},\theta)$. Now, using Equation~\ref{eq:predictive-distribution}, multi-modal prompt learning corresponds to \gls{map} estimation:
\begin{align}
    \theta_{\text{MAP}}^{\star} = \text{arg max}_{\theta}\text{ }\log p(\theta) + \sum_{i=1}^{n}p(y_{i}|u_{i},\theta),\label{eq:map}
\end{align}
where $\theta$ includes all learnable prompts and coupling parameters. As a consequence of Equation~\ref{eq:map}, MaPLe and MMRL inherit the intrinsic limitations of \gls{map} estimation under data scarcity. In particular, they are subject to overfitting, and do not account for predictive uncertainty. 

Our goal is to use a Bayesian setting, which allows us to sample high quality prompts from the posterior $p(\theta|\mathcal{D})$. The main insight is that the underlying prompt landscape has many equally good (in terms of training loss) prompts that have different generalization capabilities (test loss or accuracy). On the one hand, we want to generate samples that are likely under the posterior $p(\theta|\mathcal{D})$. On the other hand, we want to enhance sample diversity to uncover multiple modes in the landscape. To achieve \textbf{both} of these goals, we propose the \gls{rcsgmcmc} algorithm. Our insight is that, in addition to using the cyclical \gls{sgmcmc} schedule of~\cite{zhang2019cyclical}, we can further encourage exploration with a repulsion term between prompts.

\begin{remark}{(Notation)}
    In the following, we describe our method mathematically. We use $k$ to denote different samples in MCMC, $t$ to denote the iteration, and $c$ to denote cycles. In this sense, each cycle $c=1,\cdots,C$ is composed of $t=1,\cdots,T$ iterations across the $k=1,\cdots,K$ samples.
\end{remark}
Given a balance parameter $\beta$, we define the \textcolor{cvprblue}{exploration stage} where $\frac{t}{T} \le \beta$, and the \textcolor{cvprred}{sampling stage} where $\frac{t}{T} > \beta$. We sample using Equation~\ref{eq:sampling} :

\begin{equation}
    \begin{cases}
        \theta_{k,t+1}^{(c)} = \theta_{k,t}^{(c)} + r_{k,t}^{(c)} + \textcolor{cvprgreen}{\xi\sum_{\ell=1}^{K}F(\theta_{k,t}^{(c)}, \theta_{\ell, T}^{(c-1)})},\\
        \begin{aligned}
        r_{k,t+1}^{(c)} = (1-\eta)r_{k,t}^{(c)} &- \textcolor{cvprblue}{\alpha_{t}\nabla \tilde{U}(\theta_{k,t}^{(c)})} \\
        &+ \textcolor{cvprred}{\mathbb{I}_{\frac{t}{T}, \beta}\sqrt{2(\eta - \hat{\gamma})\alpha_{t}}\epsilon_{t}},
        \end{aligned}
    \end{cases}\label{eq:sampling}
\end{equation}

for $\epsilon_{t} \sim \mathcal{N}(0, \text{I}_{d})$, and $\mathbb{I}_{\frac{t}{T}, \beta}=1$ if $\frac{t}{T}> \beta$ and 0 otherwise. These two stages are scheduled within cycles, as in~\citet{zhang2019cyclical}. We provide further mathematical motivation for these equations in our supplementary materials.

The main feature that distinguishes Equation~\ref{eq:sampling} is the repulsion of samples from the current cycle $c$, $\{\theta_{k,t}^{(c)}\}_{k=1}^{K}$, away from those of the previous cycle, $\{\theta_{\ell,T}^{(c-1)}\}_{\ell=1}^{K}$. We refer readers to Figure~\ref{fig:illustration} for a conceptual illustration of the effect of this force on the \gls{sghmc} algorithm, i.e., mode exploration in the posterior.


We model the \textcolor{cvprgreen}{repulsive force} through another potential, $V(\theta,\theta')$, so that, $F(\theta, \theta') = -\nabla_{\theta} V(\theta, \theta')$. Intuitively, this potential should be large for similar parameters, and small for different ones. A natural potential is,
\begin{align}
    V(\theta,\theta') &= \dfrac{1}{d_{\Theta}(\theta,\theta')^{2}+\epsilon},\label{eq:potential}
\end{align}
where $d_{\Theta}:\Theta^{2}\rightarrow\mathbb{R}$ is a distance in the space of parameters $\Theta$. We propose comparing the representations extracted by the networks, i.e., $u_{\theta,i}$ for a representation of image $x_i$;
 Algorithm~\ref{alg:csghmc} gives our complete repulsive cSGHMC approach.


Modeling the geometry of the weight space is challenging~\cite{li2018visualizing} due to invariance to permutations (e.g.~\cite{gelberg2024variational}) and data scarcity in this space. Therefore, we propose comparing parameters $\theta$ and $\theta'$ in terms of the distribution of their activations, i.e.,
\begin{align}
    d_{\Theta}(\theta,\theta') = d_{\mathcal{P}(\mathcal{U})}(U_{\theta}, U_{\theta'}),\label{eq:dist-params}
\end{align}
where $U_{\theta} = \{ u_{\theta,i} \}_{i=1}^{n}$ and $d_{\mathcal{P}(\mathcal{U})}$ is a distance over the set of probability distributions over $\mathcal{U}$. The MMD~\cite{gretton2012kernel} and Wasserstein distance~\cite{montesuma2024recent} are of main interest to our applications, which are described in Section~\ref{sec:prob-metrics}.

While computing the \gls{mmd} or the Wasserstein distance incurs an additional computational overhead, Equation~\ref{eq:dist-params} is computed on the level of mini-batches (e.g., 32 samples). Since the complexity of these distances scales with the number of samples (i.e., $\mathcal{O}(n^{2})$ and $\mathcal{O}(n^{3})$ respectively), the overall computational cost of $d_{\mathcal{P}(\mathcal{U})}$ is negligible (c.f. Appendix~\ref{appx:overhead}).



\begin{algorithm}[t]
\caption{Repulsive cSGHMC Training with Cycle Restarts and Inter-Cycle Repulsion}
\label{alg:csghmc}
\begin{algorithmic}[1]
\Require Initial step-size $\alpha_{0}$, number of iterations $T$, number of cycles $C$, proportion of exploration $\beta$, momentum $1 - \eta$, repulsion strength $\xi$, noise estimate $\hat{\gamma}$. 
\For{Cycle $c=1,\cdots,C$}
    \For{Iteration $t=1,\cdots,T$}
        \State Set $\alpha_{t}$ with cosine scheduling
        \State $F_{k,t}^{(c)} = \begin{cases}
            \sum_{\ell=1}^{K}F(\theta_{k,t}^{(c)}, \theta_{\ell,T}^{(c-1)})&c>1\\
            0&\text{otherwise}
        \end{cases}$
        \State $n_{t} = \begin{cases}
            \sqrt{2(\eta - \hat{\gamma})\alpha_{t}}\epsilon_{t} & \frac{\text{mod}(t - 1, \lceil T / C \rceil)}{\lceil T / C \rceil} > \beta\\
            0 & \text{otherwise}
        \end{cases}$
        \State $\theta_{k,t+1}^{(c)} = \theta_{k,t}^{(c)} + r_{k,t}^{(c)} + F_{k,t}^{(c)},$
        \State $r_{k,t+1}^{(c)} = (1-\eta)r_{k,t}^{(c)} - \alpha_{t}\nabla \tilde{U}(\theta_{k,t}^{(c)}) + n_{t}$
    \EndFor
\EndFor
\end{algorithmic}
\end{algorithm}

After the execution of Algorithm~\ref{alg:csghmc}, we produce a set of network parameter samples $\{ \{\theta_{k,T}^{(c)}\}_{k=1}^{K} \}_{c=1}^{C}$, which includes prompts and parameters of the coupling function. As a result, we compute predictions based on the ensembling of these samples, namely,
\begin{align*}
    p(y|x) = \sum_{c=1}^{C}\sum_{k=1}^{K}\omega_{c,k}p(y|x,\theta_{k,T}^{(c)}),
\end{align*}
where $\omega_{c,k}$ is the importance of each $\theta_{k,T}^{(c)}$. For simplicity, we use uniform weighting $\omega_{c,k} = (C\cdot K)^{-1}$. Running inference on the $C \cdot K$ models is embarrassingly parallelizable (c.f. Appendix~\ref{appx:efficiency}) and adds a small computational overhead.

\definecolor{highlight}{RGB}{255,230,220}

\section{Experiments}




\noindent\textbf{Overview.} We assess the effectiveness of our proposed approach across three distinct evaluation protocols: base-to-novel class generalization, domain generalization, and cross-dataset transfer. Following established protocols in recent work on prompt learning~\cite{zhou2022learning,zhou2022conditional,khattak2023maple,guo2025mmrl}, all experiments are conducted under a 16-shot learning scenario, where we are provided with only 16 labeled training samples per category. We run experiments on ReBaPL-based extensions of the MaPLe and MMRL methods for multi-modal prompt learning. Full details on our experiments are available in the supplementary material (Appendix~\ref{appx:additional-experiments}).

\begin{table*}[ht]
\scriptsize
\centering
\caption{\textbf{Base to novel generalization results}. Overall, our ReBaPL method improves the base performance of both MaPLe and MMRL, with a considerable increase on the FGVCAircraft and EuroSAT datasets. We used MMD for the MaPLe + ReBaPL method, and Wasserstein distance for the MMRL + ReBaPL method, which respectively performed best. An asterisk (*) denotes methods we re-ran using the same random seeds and hardware. Delta values ($\Delta$) show improvements from adding ReBaPL ({\color{green!70!black}green} for positive, {\color{red}red} for negative).}
\label{tab:main_results}
\resizebox{\linewidth}{!}{
\begin{tabular}{l|ccc|ccc|ccc|ccc}
\hline
Method & \multicolumn{3}{c|}{Average} & \multicolumn{3}{c|}{ImageNet} & \multicolumn{3}{c|}{Caltech101} & \multicolumn{3}{c}{OxfordPets} \\
 & Base & Novel & HM & Base & Novel & HM & Base & Novel & HM & Base & Novel & HM \\
\hline
CLIP~\cite{radford2021learning} & 69.34 & 74.22 & 71.70 & 72.43 & 68.14 & 70.22 & 96.84 & 94.00 & 95.40 & 91.17 & 97.26 & 94.12 \\
CoOp~\cite{zhou2022learning} & 82.69 & 63.22 & 70.83 & 76.47 & 67.88 & 71.92 & 98.00 & 89.81 & 93.73 & 93.67 & 95.29 & 94.47 \\
CoCoOp~\cite{zhou2022learning} & 80.47 & 71.69 & 75.83 & 75.98 & 70.43 & 73.10 & 97.96 & 93.81 & 95.84 & 95.20 & 97.69 & 96.43 \\
APP~\cite{cho2024make} &  83.0 & 65.8 & 72.61 & 69.9 & 63.2 & 66.4 & 95.2 & 91.0 & 93.0 & 96.8 & 88.3 & 92.4 \\

PromptSRC*~\cite{khattak2023self} & 84.93 & 74.49 & 78.61 & 76.77 & 67.8 & 72.01 & 98.07 & 94.03 & 96.01 & 95.27 & 97.23 & 96.24 \\
\hline
MaPLe*~\cite{khattak2023maple} & 82.03 & 75.03 & 78.37 & 74.96 & 66.97 & 70.74 & 97.83 & 94.87 & 96.33 & 95.20 & 98.13 & 96.64\\
\rowcolor{red!15}
MaPLe* + ReBaPL & 83.28 & 76.08 & 79.52 & 76.06 & 68.80 & 72.25 & 98.35 & 94.87 & 96.58 & \textbf{96.17} & \textbf{97.77} & \textbf{96.96} \\

\multicolumn{1}{c}{$\Delta$} & {\textcolor{green!70!black}{\scriptsize +1.25}} & {\textcolor{green!70!black}{\scriptsize +1.05}} & {\textcolor{green!70!black}{\scriptsize +1.15}} & {\textcolor{green!70!black}{\scriptsize +1.10}} & {\textcolor{green!70!black}{\scriptsize +1.83}} & {\textcolor{green!70!black}{\scriptsize +1.51}} & {\textcolor{green!70!black}{\scriptsize +0.52}} & {\scriptsize 0.0} & {\textcolor{green!70!black}{\scriptsize +0.25}} & {\textcolor{green!70!black}{\scriptsize +0.97}} & {\textcolor{red}{\scriptsize -0.36}} & {\textcolor{green!70!black}{\scriptsize +0.32}} \\

MMRL*~\cite{guo2025mmrl} & {85.54} & 76.52 & 80.59 & 77.55 & 67.43 & 72.14 & 98.93 & \textbf{94.60} & \textbf{96.72} & 95.27 & 97.23 & 96.24\\
\rowcolor{red!15}

MMRL* + ReBaPL & \textbf{85.74} & \textbf{77.44} & \textbf{81.38} & \textbf{77.90} & \textbf{68.83} & \textbf{73.09} & \textbf{99.10} & 94.27 & 96.62 & 95.93 & 97.57 & 96.74\\
\multicolumn{1}{c}{$\Delta$} & {\textcolor{green!70!black}{\scriptsize +0.20}} & {\textcolor{green!70!black}{\scriptsize +0.92}} & {\textcolor{green!70!black}{\scriptsize +0.79}} & {\textcolor{green!70!black}{\scriptsize +0.35}} & {\textcolor{green!70!black}{\scriptsize +1.40}} & {\textcolor{green!70!black}{\scriptsize +0.95}} & {\textcolor{green!70!black}{\scriptsize +0.17}} & {\textcolor{red}{\scriptsize -0.33}} & {\textcolor{red}{-0.10}} & {\textcolor{green!70!black}{\scriptsize +0.66}} & {\textcolor{green!70!black}{\scriptsize +0.34}} & {\textcolor{green!70!black}{\scriptsize +0.50}} \\

\hline
\hline
Method & \multicolumn{3}{c|}{StanfordCars} & \multicolumn{3}{c|}{Flowers102} & \multicolumn{3}{c|}{Food101} & \multicolumn{3}{c}{FGVCAircraft} \\
 & Base & Novel & HM & Base & Novel & HM & Base & Novel & HM & Base & Novel & HM \\
\hline
CLIP~\cite{radford2021learning} & 63.37 & 74.89 & 68.65 & 72.08 & 77.80 & 74.83 & 90.10 & 91.22 & 90.66 & 27.19 & 36.29 & 31.09 \\
CoOp~\cite{zhou2022learning} & 78.12 & 60.40 & 68.13 & 97.60 & 59.67 & 74.06 & 88.33 & 82.26 & 85.19 & 40.44 & 22.30 & 28.75 \\
CoCoOp~\cite{zhou2022learning} & 70.49 & 73.59 & 72.01 & 94.87 & 71.75 & 81.71 & 90.70 & 91.29 & 90.99 & 33.41 & 23.71 & 27.74 \\
APP~\cite{cho2024make} &  85.9 & 69.5 & 76.8 & 96.8 & 61.0 & 74.8 & 84.6 & 86.1 & 85.4 & 44.9 & 26.0 & 33.0 \\
PromptSRC*~\cite{khattak2023self} & 77.93 & 75.5 & 76.7 & 97.83 & 77.13 & 86.26 & 90.60 & 91.53 & 91.06 & 41.43 & 23.67 & 30.13 \\
\hline
MaPLe*~\cite{khattak2023maple} & 72.45 & 74.90 & 73.65 & 96.33 & 73.33 & 83.27 & 90.80 & 92.10 & 91.45 & 36.60 & 34.90 & 35.73\\
\rowcolor{red!15}

MaPLe* + ReBaPL & 74.73 & 74.57 & 74.65 & 97.43 & 74.37 & 84.35 & \textbf{90.83} & \textbf{92.13} & \textbf{91.48} & 38.00 & 34.03 & 35.91\\

\multicolumn{1}{c}{$\Delta$} & {\textcolor{green!70!black}{\scriptsize +2.28}} & {\textcolor{red}{\scriptsize -0.33}} & {\textcolor{green!70!black}{\scriptsize +1.0}} & {\textcolor{green!70!black}{\scriptsize +1.10}} & {\textcolor{green!70!black}{\scriptsize +1.04}} & {\textcolor{green!70!black}{\scriptsize +1.08}} & {\textcolor{green!70!black}{\scriptsize +0.03}} & {\textcolor{green!70!black}{\scriptsize +0.03}} & {\textcolor{green!70!black}{\scriptsize +0.03}} & {\textcolor{green!70!black}{\scriptsize +1.4}} & {\textcolor{red}{\scriptsize -0.87}} & {\textcolor{green!70!black}{\scriptsize +0.18}} \\

MMRL*~\cite{guo2025mmrl} & \textbf{81.23} & 75.00 & 77.99 & 98.70 & 76.83 & 86.40 & 90.60 & 91.53 & 91.06 & \textbf{45.70} & 37.1 & 40.95\\
\rowcolor{red!15}

MMRL* + ReBaPL & 81.20 & \textbf{75.37} & \textbf{78.17} & \textbf{98.80} & \textbf{77.23} & \textbf{86.70} & 90.73 & 91.60 & 91.16 & {45.13} & \textbf{38.57} & \textbf{41.59}\\
\multicolumn{1}{c}{$\Delta$} & {\textcolor{red}{-0.03}} & {\textcolor{green!70!black}{+0.37}} & {\textcolor{green!70!black}{\scriptsize +0.18}} & {\textcolor{green!70!black}{\scriptsize +0.10}} & {\textcolor{green!70!black}{\scriptsize +0.40}} & {\textcolor{green!70!black}{\scriptsize +0.30}} & {\textcolor{green!70!black}{\scriptsize +0.13}} & {\textcolor{green!70!black}{\scriptsize +0.07}} & {\textcolor{green!70!black}{\scriptsize +0.10}} & {\textcolor{red}{\scriptsize -0.57}} & {\textcolor{green!70!black}{\scriptsize +1.47}} & {\textcolor{green!70!black}{\scriptsize +0.64}} \\

\hline
\hline
Method & \multicolumn{3}{c|}{SUN397} & \multicolumn{3}{c|}{DTD} & \multicolumn{3}{c|}{EuroSAT} & \multicolumn{3}{c}{UCF101} \\
 & Base & Novel & HM & Base & Novel & HM & Base & Novel & HM & Base & Novel & HM \\
\hline
CLIP~\cite{radford2021learning} & 69.36 & 75.35 & 72.23 & 53.24 & 59.90 & 56.37 & 56.48 & 64.05 & 60.03 & 70.53 & 77.50 & 73.85 \\
CoOp~\cite{zhou2022learning} & 80.60 & 65.89 & 72.51 & 79.44 & 41.18 & 54.24 & 92.19 & 54.74 & 68.69 & 84.69 & 56.05 & 67.46 \\
CoCoOp~\cite{zhou2022learning} & 79.74 & 76.86 & 78.27 & 77.01 & 56.00 & 64.85 & 87.49 & 60.04 & 71.21 & 82.33 & 73.45 & 77.64 \\
APP~\cite{cho2024make} & 80.6 & 73.3 & 76.8 & 78.4 & 48.9 & 60.2 & 93.6 & 47.6 & 63.1 & 86.2 & 69.2 & 76.8 \\
PromptSRC*~\cite{khattak2023self} & 92.93 & 74.17 & 80.53 & 83.33 & 61.3 & 70.64 & 92.93 & 74.17 & 82.5 & 87.10 & 78.53 & 82.6 \\
\hline
MaPLe*~\cite{khattak2023maple} & 81.07 & 77.90 & 79.45 & 81.30 & 57.50 & 67.36 & 92.47 & 77.03 & 84.05 & 83.97 & 77.70 & 80.71\\
\rowcolor{red!15}
MaPLe* + ReBaPL & 81.83 & \textbf{79.87} & 80.84 &  82.50 & 61.60 & 70.53 & 95.30 & 79.20 & 86.51 & 84.97 & 79.73 & 82.27 \\

\multicolumn{1}{c}{$\Delta$} & {\textcolor{green!70!black}{\scriptsize +0.76}} & {\textcolor{green!70!black}{\scriptsize +1.97}} & {\textcolor{green!70!black}{\scriptsize +1.39}} & {\textcolor{green!70!black}{\scriptsize +1.2}} & {\textcolor{green!70!black}{\scriptsize +4.10}} & {\textcolor{green!70!black}{\scriptsize +3.17}} & {\textcolor{green!70!black}{\scriptsize +2.83}} & {\textcolor{green!70!black}{\scriptsize +2.17}} & {\textcolor{green!70!black}{\scriptsize +2.46}} & {\textcolor{green!70!black}{\scriptsize +1.0}} & {\textcolor{green!70!black}{\scriptsize +2.03}} & {\textcolor{green!70!black}{\scriptsize +1.56}} \\

MMRL*~\cite{guo2025mmrl} & 83.23 & 79.27 & 81.20 & 85.63 & \textbf{65.13} & 73.99 & 95.80 & 77.20 & 85.50 & 88.30 & 79.70 & 83.78\\
\rowcolor{red!15}
MMRL* + ReBaPL & \textbf{83.20} & 79.37 & \textbf{81.24} & \textbf{86.10} & 65.00 & \textbf{74.08} & \textbf{96.73} & \textbf{83.63} & \textbf{89.71} & \textbf{88.33} & \textbf{80.40} & \textbf{84.18}\\
\multicolumn{1}{c}{$\Delta$} & {\textcolor{red}{-0.03}} & {\textcolor{green!70!black}{+0.10}} & {\textcolor{green!70!black}{\scriptsize +0.04}} & {\textcolor{green!70!black}{\scriptsize +0.47}} & {\textcolor{red}{\scriptsize -0.13}} & {\textcolor{green!70!black}{\scriptsize +0.09}} & {\textcolor{green!70!black}{\scriptsize +0.93}} & {\textcolor{green!70!black}{\scriptsize +6.43}} & {\textcolor{green!70!black}{\scriptsize +4.21}} & {\textcolor{green!70!black}{\scriptsize +0.03}} & {\textcolor{green!70!black}{\scriptsize +0.70}} & {\textcolor{green!70!black}{\scriptsize +0.40}} \\

\hline
\end{tabular}
}
\end{table*}

\subsection{Base to Novel Generalization}

In this protocol, the available classes within each dataset are partitioned into two disjoint subsets: base categories and novel categories. During training, the model has access exclusively to labeled examples from the base categories. Evaluation is then performed on both base and novel subsets, enabling us to measure both the model's adaptation performance on seen classes and its capacity to preserve zero-shot generalization on previously unseen classes. This evaluation is carried out across 11 benchmark classification datasets: ImageNet~\cite{deng2009imagenet}, Caltech101~\cite{fei2004caltech101}, OxfordPets~\cite{parkhi2012cats}, StanfordCars~\cite{krause20133d}, Flowers102~\cite{nilsback2008automated}, Food101~\cite{bossard2014food}, FGVCAircraft~\cite{ori2013fine}, SUN397~\cite{xiao2010sun}, UCF101~\cite{soomro2012ucf101}, DTD~\cite{cimpoi2014describing}, and EuroSAT~\cite{helber2019eurosat}.

In Table \ref{tab:main_results} we present the performance of our method in the base-to-novel generalization task. We compare our method to the CLIP~\cite{radford2021learning} baseline, and other methods including CoOp~\cite{zhou2022learning}, and CoCoOp~\cite{zhou2022conditional}, APP~\cite{cho2024make}, PromptSRC~\cite{khattak2023self}, MaPLe~\cite{khattak2023maple} and MMRL~\cite{guo2025mmrl}. We re-run some of the leading methods (PromptSRC, MaPLe, and MMRL) to ensure a fair comparison and avoid inconsistent results due to different hardware. While VaMP~\cite{chengvamp2025} is a method related to ours, its code is not publicly available, and thus we do not include it our experiments. We demonstrate the effectiveness of our proposed ReBaPL approach by running it as extensions of MaPLe and MMRL. 

In comparison with MaPLe, our method consistently improves performance on base classes, and in many cases also  novel classes. For some datasets (e.g., OxfordPets, StanfordCars, and FGVCAircraft), MaPLe outperforms our method by a small margin. Overall, we consistently improve over MaPLe in terms of the harmonic mean of base and novel accuracy, striking a balance in generalization.

Concerning MMRL, while we improve on both base and novel classes, our gains are mostly on the novel classes. This finding highlights our claim that our repulsive cyclical SGHMC algorithm, and hence ReBaPL, improves generalization by exploring the posterior landscape more thoroughly. On average, our method improves MMRL on both base and novel classes, with a larger margin on the novel classes. Overall the harmonic mean is also improved. Our method has substantial gains on the EuroSAT and ImageNet datasets, establishing a new state-of-the-art.


\subsection{Cross-Dataset Transfer}

To evaluate robustness under dataset shift, we adopt a source-to-target transfer protocol. The model is first trained on the complete set of 1000 ImageNet categories using the 16-shot setting, and evaluated on the remaining 10 datasets without any additional fine-tuning. This setup allows us to measure how well learned prompt adaptations transfer to entirely different data distributions and visual domains.

We see from Table~\ref{tab:cross_dataset} that our ReBaPL-based models provide significant gains over the baseline MaPLe and MMRL methods. Notably, our MMRL + ReBaPL approach provides the highest average accuracy of 67.62\%, indicating better generalization performance than competing methods.

\begin{table*}[!htbp]
\centering
\setlength{\abovecaptionskip}{0.15cm}  
\caption{Comparison of our ReBaPL method with previous state-of-the-art multi-modal prompt learning methods on cross-dataset evaluation across 10 datasets. An asterisk (*) denotes methods we re-ran using the same random seeds and hardware. Delta values ($\Delta$) show improvements from adding ReBaPL ({\color{green!70!black}green} for positive, {\color{red}red} for negative).}
\label{tab:cross_dataset}
    \scriptsize
    \begin{tabular}{@{}rc|ccccccccccc@{}}
    \toprule
     &
      \textbf{Source} &
      \multicolumn{11}{c}{\textbf{Target}} \\ \cmidrule(l){2-13} 
     &
      \rotatebox{60}{ImageNet} &
      \rotatebox{60}{Caltech101} &
      \rotatebox{60}{OxfordPets} &
      \rotatebox{60}{StanfordCars} &
      \rotatebox{60}{Flowers101} &
      \rotatebox{60}{Food101} &
      \rotatebox{60}{FGVCAircraft} &
      \rotatebox{60}{SUN397} &
      \rotatebox{60}{DTD} &
      \rotatebox{60}{EuroSAT} &
      \rotatebox{60}{UCF101} &
      \rotatebox{60}{Average} \\ \cmidrule(l){2-13}
    \\$\text{PromptSRC*}~\cite{khattak2023self}$ &
     68.96 &
      93.47&
      90.33&
      65.87&
      70.40&
      83.66&
      24.13&
      67.17&
      46.40&
      45.97&
      67.67 & 
      65.50 
      \\$\text{MaPLe*}~\cite{khattak2023maple}$ &
      67.96 &
      93.17 &
      90.20 &
      65.97 &
      71.07 &
      86.33 &
      23.23 &
      67.23 &
      47.20 &
      45.70 &
      66.27 &
      65.63 
      \\
\rowcolor{red!15}
    $\text{MaPLe + ReBaPL}$ &
     68.66 &
     93.80 &
     90.73 &
     66.30 &
     72.57 &
     86.40 &
     24.37  &
     67.70  &
     47.17  &
     50.90 &
     67.87 &
     66.77 \\
    \textcolor{gray}{ $\Delta$} &
     \textcolor{green!70!black}{ +0.70} &
     \textcolor{green!70!black}{ +0.63} &
     \textcolor{green!70!black}{ +0.53} &
     \textcolor{green!70!black}{ +0.33} &
     \textcolor{green!70!black}{ +1.50} &
     \textcolor{green!70!black}{ +0.07} &
     \textcolor{green!70!black}{ +1.14} &
     \textcolor{green!70!black}{ +0.47} &
     \textcolor{red}{ -0.03} &
     \textcolor{green!70!black}{ +5.20} &
     \textcolor{green!70!black}{ +1.60} &
     \textcolor{green!70!black}{ +1.14} \\
    $\text{MMRL*}~\cite{guo2025mmrl}$ &
      70.13 &
      94.30 &
      91.0 &
      66.20 &
      71.53 &
      86.20 &
      26.07 &
      67.50 &
      46.93 &
      49.83 &
      69.13 &
      66.87 \\
\rowcolor{red!15}
    $\text{MMRL + ReBaPL}$ &
      \textbf{71.0} &
      \textbf{94.60} &
      \textbf{91.97} &
      \textbf{66.63} &
      \textbf{72.57} &
      \textbf{86.43} &
      \textbf{26.33} &
      \textbf{67.90} &
      \textbf{47.27} &
      \textbf{52.97} &
      \textbf{69.50} &
      \textbf{67.62}\\
    \textcolor{gray}{ $\Delta$} &
     \textcolor{green!70!black}{ +0.87} &
     \textcolor{green!70!black}{ +0.30} &
     \textcolor{green!70!black}{ +0.97} &
     \textcolor{green!70!black}{ +0.43} &
     \textcolor{green!70!black}{ +1.04} &
     \textcolor{green!70!black}{ +0.23} &
     \textcolor{green!70!black}{ +0.26} &
     \textcolor{green!70!black}{ +0.40} &
     \textcolor{green!70!black}{ +0.34} &
     \textcolor{green!70!black}{ +3.14} &
     \textcolor{green!70!black}{ +0.37} &
     \textcolor{green!70!black}{ +0.75} \\
      
     \bottomrule
    \end{tabular}
    
\end{table*}
\subsection{Domain Generalization} 
To assess robustness against domain shift and out-of-distribution scenarios, we train exclusively on ImageNet and evaluate on four domain-shifted variants: ImageNetV2~\cite{recht2019imagenet}, ImageNet-Sketch~\cite{wang2019learning}, ImageNet-A~\cite{hendrycks2021natural}, and ImageNet-R~\cite{hendrycks2021many}. 

We show in Table~\ref{tab:domaingeneralization} that our ReBaPL methods improve generalization on out-of-domain datasets, compared to the underlying MLE-based prompt learning method. We also see that our MaPLe + ReBaPL method provides the best performance on the ImageNet-A target. Finally, we see that our MMRL + ReBaPL method provides the highest accuracy on the source ImageNet dataset. Taken together, these results show that our ReBaPL approach is effective at improving out-of-domain generalization performance without hindering the performance on the source domain.
\begin{table}[t]
\centering
\setlength{\tabcolsep}{3.0pt}
\resizebox{\columnwidth}{!}{%
    \begin{tabular}{l c | c c c c}
    \toprule
    & \multicolumn{1}{c|}{\textbf{Source}} & \multicolumn{3}{c}{\textbf{Target}} & \\
    \cmidrule(lr){2-6}
    Method & ImageNet & ImageNetV2 & ImageNet-S & ImageNet-A & ImageNet-R \\
    \midrule
    $\text{PromptSRC*}~\cite{khattak2023self}$  & 68.96 & 62.50 & \textbf{48.60} & 49.63 & \textbf{75.77} \\ 
    $\text{MaPLe*}~\cite{khattak2023maple}$ & 67.96 & 61.57 & 47.70 & 48.80 & 75.33 \\ 
    \rowcolor{red!15}
    MaPLe + ReBaPL   & 68.66 & 62.30 & 48.50 & \textbf{49.73} & 75.40 \\ 
    \textcolor{gray}{ $\Delta$} & 
    \textcolor{green!70!black}{ +0.70} & 
    \textcolor{green!70!black}{ +0.73} & 
    \textcolor{green!70!black}{ +0.80} & 
    \textcolor{green!70!black}{ +0.93} & 
    \textcolor{green!70!black}{ +0.07} \\ 
    $\text{MMRL*}~\cite{guo2025mmrl}$  & 70.13 & 62.20 & 47.80 & 48.90 & 75.03 \\ 
    \rowcolor{red!15}
    MMRL + ReBaPL    & \textbf{71.00} & \textbf{62.50} & 48.40 & 49.63 & 75.63 \\ 
    \textcolor{gray}{ $\Delta$} & 
    \textcolor{green!70!black}{ +0.87} & 
    \textcolor{green!70!black}{ +0.30} & 
    \textcolor{green!70!black}{ +0.60} & 
    \textcolor{green!70!black}{ +0.73} & 
    \textcolor{green!70!black}{ +0.60} \\ 
    \bottomrule
    \end{tabular}%
}
\caption{Comparison of robustness on out-of-distribution datasets. Delta values ($\Delta$) show improvements from adding ReBaPL ({\color{green!70!black}green} for positive, {\color{red}red} for negative).}
\label{tab:domaingeneralization}
\end{table}

\subsection{Ablation}

We ablate a few design choices in our method. We compare 3 main choices: ReBaPL without repulsion (i.e., $F(\theta,\theta') := 0$) vs. ReBaPL with repulsion based on the Wasserstein distance and the MMD. This ablation seeks to isolate the benefit of using repulsion, and demonstrate that our method is robust to the choice of probability metric. We evaluate our methods on all datasets listed in Table~\ref{tab:main_results}.

\begin{table}[t]
    \footnotesize
    \centering
    \caption{Ablation on the use of repulsion in ReBaPL, averaged over 11 datasets.} 
    \begin{tabular}{lccc}
        \toprule
         Method & Base & Novel & HM \\
         \midrule
         MaPLe & 82.03 & 75.03 & 78.37 \\ 
         \quad+ ReBaPL (No Repulsion) & 83.39 & 75.47 & 78.93\\
         \quad+ ReBaPL (Wasserstein) & 83.39 &75.86 & 79.44 \\
         \quad+ ReBaPL (MMD) & 83.28 & \textbf{76.08} & \textbf{79.52} \\
         \bottomrule
    \end{tabular}
    \label{tab:ablation}
\end{table}


Overall, as shown in Table~\ref{tab:ablation}, our method has stable performance for both the Wasserstein distance and MMD. The gap in harmonic mean between these choices is $0.08\%$, which is marginal compared to the improvement over MaPLe (around $1\%$). Furthermore, even without repulsion, our ReBaPL method improves over MaPLe, highlighting the benefits of sampling from the posterior distribution. By adding the repulsion term we improve performance for novel classes. This finding supports our claim that adding repulsion allows us to explore the posterior landscape more thoroughly, and thus improve generalization.

\section{Conclusion}

In this paper we have introduced Repulsive Bayesian Prompt Learning (ReBaPL), a novel approach that addresses overfitting and out-of-distribution generalization challenges in prompt learning methods. Our key contribution is the repulsive cyclical SGHMC (rcSGHMC) algorithm, which leverages Hamiltonian dynamics with cyclical learning rate schedules to alternate between exploration and exploitation phases when sampling from complex multimodal posterior distributions. By introducing a representation-based repulsive force derived from probability metrics (MMD and Wasserstein distance) computed on representation distributions, our method captures functional similarity between prompts and encourages diverse mode discovery while preventing premature convergence. Unlike prior Bayesian prompt learning approaches that rely on restrictive unimodal approximations or deterministic variational methods, ReBaPL provides a modular, plug-and-play framework that can extend any existing MLE-based prompt learning method with principled Bayesian inference. Through comprehensive experiments we have demonstrated superior generalization by maintaining a richer characterization of the prompt posterior landscape. Future work includes exploring alternative probability metrics beyond MMD and Wasserstein distance, such as Sinkhorn divergence or information-theoretic measures, and developing adaptive cyclical mechanisms that automatically adjust the number of cycles based on convergence diagnostics or posterior diversity to improve efficiency and performance.
{
    \small
    \bibliographystyle{ieeenat_fullname}
    \bibliography{main}

\begin{thebibliography}{48}
\providecommand{\natexlab}[1]{#1}
\providecommand{\url}[1]{\texttt{#1}}
\expandafter\ifx\csname urlstyle\endcsname\relax
  \providecommand{\doi}[1]{doi: #1}\else
  \providecommand{\doi}{doi: \begingroup \urlstyle{rm}\Url}\fi

\bibitem[Bendou et~al.(2025)Bendou, Ouasfi, Gripon, and Boukhayma]{proker}
Yassir Bendou, Amine Ouasfi, Vincent Gripon, and Adnane Boukhayma.
\newblock Proker: A kernel perspective on few-shot adaptation of large vision-language models.
\newblock In \emph{Proceedings of the Computer Vision and Pattern Recognition Conference}, pages 25092--25102, 2025.

\bibitem[Bossard et~al.(2014)Bossard, Guillaumin, and Van~Gool]{bossard2014food}
Lukas Bossard, Matthieu Guillaumin, and Luc Van~Gool.
\newblock {Food-101--mining discriminative components with random forests}.
\newblock In \emph{European Conference on Computer Vision}, pages 446--461. Springer, 2014.

\bibitem[Chen et~al.(2014)Chen, Fox, and Guestrin]{sghmc}
Tianqi Chen, Emily Fox, and Carlos Guestrin.
\newblock Stochastic gradient hamiltonian monte carlo.
\newblock In \emph{Proceedings of the 31st International Conference on Machine Learning}, pages 1683--1691, 2014.

\bibitem[Cheng and Han()]{chengvamp2025}
Silin Cheng and Kai Han.
\newblock Vamp: Variational multi-modal prompt learning for vision-language models.
\newblock In \emph{NeurIPS 2025}.

\bibitem[Cho et~al.(2024)Cho, Bae, Shin, Youn, Joo, and Moon]{cho2024make}
Youngjae Cho, HeeSun Bae, Seungjae Shin, Yeo~Dong Youn, Weonyoung Joo, and Il-Chul Moon.
\newblock Make prompts adaptable: Bayesian modeling for vision-language prompt learning with data-dependent prior.
\newblock In \emph{AAAI}, pages 11552--11560, 2024.

\bibitem[Cimpoi et~al.(2014)Cimpoi, Maji, Kokkinos, Mohamed, and Vedaldi]{cimpoi2014describing}
Mircea Cimpoi, Subhransu Maji, Iasonas Kokkinos, Sammy Mohamed, and Andrea Vedaldi.
\newblock {Describing textures in the wild}.
\newblock In \emph{IEEE Conference on Computer Vision and Pattern Recognition}, pages 3606--3613, 2014.

\bibitem[Cuturi(2013)]{cuturi2013sinkhorn}
Marco Cuturi.
\newblock Sinkhorn distances: Lightspeed computation of optimal transport.
\newblock \emph{Advances in neural information processing systems}, 26, 2013.

\bibitem[Deng et~al.(2009)Deng, Dong, Socher, Li, Li, and Fei-Fei]{deng2009imagenet}
Jia Deng, Wei Dong, Richard Socher, Li-Jia Li, Kai Li, and Li Fei-Fei.
\newblock {ImageNet: A large-scale hierarchical image database}.
\newblock In \emph{IEEE Conference on Computer Vision and Pattern Recognition}, pages 248--255. IEEE, 2009.

\bibitem[Derakhshani et~al.(2023)Derakhshani, Sanchez, Bulat, da~Costa, Snoek, Tzimiropoulos, and Martinez]{derakhshani2023bayesian}
Mohammad~Mahdi Derakhshani, Enrique Sanchez, Adrian Bulat, Victor G~Turrisi da Costa, Cees~GM Snoek, Georgios Tzimiropoulos, and Brais Martinez.
\newblock Bayesian prompt learning for image-language model generalization.
\newblock In \emph{IEEE/CVF International Conference on Computer Vision}, pages 15237--15246, 2023.

\bibitem[Dosovitskiy et~al.(2020)Dosovitskiy, Beyer, Kolesnikov, Weissenborn, Zhai, Unterthiner, Dehghani, Minderer, Heigold, Gelly, et~al.]{dosovitskiy2020image}
Alexey Dosovitskiy, Lucas Beyer, Alexander Kolesnikov, Dirk Weissenborn, Xiaohua Zhai, Thomas Unterthiner, Mostafa Dehghani, Matthias Minderer, Georg Heigold, Sylvain Gelly, et~al.
\newblock An image is worth 16x16 words: Transformers for image recognition at scale.
\newblock \emph{arXiv preprint arXiv:2010.11929}, 2020.

\bibitem[Fei-Fei et~al.(2004)Fei-Fei, Fergus, and Perona]{fei2004caltech101}
Li Fei-Fei, Rob Fergus, and Pietro Perona.
\newblock {Learning generative visual models from few training examples: An incremental bayesian approach tested on 101 object categories}.
\newblock In \emph{IEEE Conference on Computer Vision and Pattern Recognition - Workshops}, page 178. IEEE, 2004.

\bibitem[Gao et~al.(2021)Gao, Geng, Zhang, Ma, Fang, Zhang, Li, and Qiao]{clipadapter}
Peng Gao, Shijie Geng, Renrui Zhang, Teli Ma, Rongyao Fang, Yongfeng Zhang, Hongsheng Li, and Yu Qiao.
\newblock Clip-adapter: Better vision-language models with feature adapters.
\newblock \emph{arXiv preprint arXiv:2110.04544}, 2021.

\bibitem[Gelberg et~al.(2024)Gelberg, van~der Ouderaa, van~der Wilk, and Gal]{gelberg2024variational}
Yoav Gelberg, Tycho~FA van~der Ouderaa, Mark van~der Wilk, and Yarin Gal.
\newblock Variational inference failures under model symmetries: Permutation invariant posteriors for bayesian neural networks.
\newblock \emph{arXiv preprint arXiv:2408.05496}, 2024.

\bibitem[Genevay et~al.(2018)Genevay, Peyr{\'e}, and Cuturi]{genevay2018learning}
Aude Genevay, Gabriel Peyr{\'e}, and Marco Cuturi.
\newblock Learning generative models with sinkhorn divergences.
\newblock In \emph{International Conference on Artificial Intelligence and Statistics}, pages 1608--1617. PMLR, 2018.

\bibitem[Gretton et~al.(2012)Gretton, Borgwardt, Rasch, Sch{\"o}lkopf, and Smola]{gretton2012kernel}
Arthur Gretton, Karsten~M Borgwardt, Malte~J Rasch, Bernhard Sch{\"o}lkopf, and Alexander Smola.
\newblock A kernel two-sample test.
\newblock \emph{The journal of machine learning research}, 13\penalty0 (1):\penalty0 723--773, 2012.

\bibitem[Guo and Gu(2025)]{guo2025mmrl}
Yuncheng Guo and Xiaodong Gu.
\newblock Mmrl: Multi-modal representation learning for vision-language models.
\newblock In \emph{Proceedings of the Computer Vision and Pattern Recognition Conference}, pages 25015--25025, 2025.

\bibitem[He et~al.(2016)He, Zhang, Ren, and Sun]{he2016deep}
Kaiming He, Xiangyu Zhang, Shaoqing Ren, and Jian Sun.
\newblock Deep residual learning for image recognition.
\newblock In \emph{Proceedings of the IEEE conference on computer vision and pattern recognition}, pages 770--778, 2016.

\bibitem[Helber et~al.(2019)Helber, Bischke, Dengel, and Borth]{helber2019eurosat}
Patrick Helber, Benjamin Bischke, Andreas Dengel, and Damian Borth.
\newblock {Eurosat: A novel dataset and deep learning benchmark for land use and land cover classification}.
\newblock \emph{IEEE Journal of Selected Topics in Applied Earth Observations and Remote Sensing}, 12\penalty0 (7):\penalty0 2217--2226, 2019.

\bibitem[Hendrycks et~al.(2021{\natexlab{a}})Hendrycks, Basart, Mu, Kadavath, Wang, Dorundo, Desai, Zhu, Parajuli, Guo, et~al.]{hendrycks2021many}
Dan Hendrycks, Steven Basart, Norman Mu, Saurav Kadavath, Frank Wang, Evan Dorundo, Rahul Desai, Tyler Zhu, Samyak Parajuli, Mike Guo, et~al.
\newblock {The many faces of robustness: A critical analysis of out-of-distribution generalization}.
\newblock In \emph{IEEE International Conference on Computer Vision}, pages 8320--8329, 2021{\natexlab{a}}.

\bibitem[Hendrycks et~al.(2021{\natexlab{b}})Hendrycks, Zhao, Basart, Steinhardt, and Song]{hendrycks2021natural}
Dan Hendrycks, Kevin Zhao, Steven Basart, Jacob Steinhardt, and Dawn Song.
\newblock {Natural adversarial examples}.
\newblock In \emph{IEEE Conference on Computer Vision and Pattern Recognition}, pages 15262--15271, 2021{\natexlab{b}}.

\bibitem[Khattak et~al.(2023{\natexlab{a}})Khattak, Rasheed, Maaz, Khan, and Khan]{khattak2023maple}
Muhammad~Uzair Khattak, Hanoona Rasheed, Muhammad Maaz, Salman Khan, and Fahad~Shahbaz Khan.
\newblock Maple: Multi-modal prompt learning.
\newblock In \emph{Proceedings of the IEEE/CVF conference on computer vision and pattern recognition}, pages 19113--19122, 2023{\natexlab{a}}.

\bibitem[Khattak et~al.(2023{\natexlab{b}})Khattak, Wasim, Naseer, Khan, Yang, and Khan]{khattak2023self}
Muhammad~Uzair Khattak, Syed~Talal Wasim, Muzammal Naseer, Salman Khan, Ming-Hsuan Yang, and Fahad~Shahbaz Khan.
\newblock Self-regulating prompts: Foundational model adaptation without forgetting.
\newblock In \emph{2023 IEEE/CVF International Conference on Computer Vision (ICCV)}, pages 15144--15154. IEEE, 2023{\natexlab{b}}.

\bibitem[Kim et~al.(2025)Kim, Ko, and Park]{kim2025bayesian}
Mingyu Kim, Jongwoo Ko, and Mijung Park.
\newblock Bayesian principles improve prompt learning in vision-language models.
\newblock In \emph{AISTATS}, 2025.

\bibitem[Krause et~al.(2013)Krause, Stark, Deng, and Fei-Fei]{krause20133d}
Jonathan Krause, Michael Stark, Jia Deng, and Li Fei-Fei.
\newblock {3D object representations for fine-grained categorization}.
\newblock In \emph{IEEE Conference on Computer Vision and Pattern Recognition - Workshops}, pages 554--561. IEEE, 2013.

\bibitem[Li et~al.(2018)Li, Xu, Taylor, Studer, and Goldstein]{li2018visualizing}
Hao Li, Zheng Xu, Gavin Taylor, Christoph Studer, and Tom Goldstein.
\newblock Visualizing the loss landscape of neural nets.
\newblock \emph{Advances in neural information processing systems}, 31, 2018.

\bibitem[Lu et~al.(2022)Lu, Liu, Zhang, Liu, and Tian]{lu2022prompt}
Yuning Lu, Jianzhuang Liu, Yonggang Zhang, Yajing Liu, and Xinmei Tian.
\newblock Prompt distribution learning.
\newblock In \emph{Proceedings of the IEEE/CVF conference on computer vision and pattern recognition}, pages 5206--5215, 2022.

\bibitem[Ma et~al.(2022)Ma, Liu, Deng, Xie, Dong, and Xu]{Ma2022-kw}
Chengcheng Ma, Yang Liu, Jiankang Deng, Lingxi Xie, Weiming Dong, and Changsheng Xu.
\newblock Understanding and mitigating overfitting in prompt tuning for vision-language models.
\newblock \emph{arXiv [cs.CV]}, 2022.

\bibitem[Maji et~al.(2013)Maji, Rahtu, Kannala, Blaschko, and Vedaldi]{ori2013fine}
Subhransu Maji, Esa Rahtu, Juho Kannala, Matthew Blaschko, and Andrea Vedaldi.
\newblock {Fine-grained visual classification of aircraft}.
\newblock \emph{arXiv preprint arXiv:1306.5151}, 2013.

\bibitem[Montesuma et~al.(2024)Montesuma, Mboula, and Souloumiac]{montesuma2024recent}
Eduardo~Fernandes Montesuma, Fred Maurice~Ngole Mboula, and Antoine Souloumiac.
\newblock Recent advances in optimal transport for machine learning.
\newblock \emph{IEEE Transactions on Pattern Analysis and Machine Intelligence}, 2024.

\bibitem[Nilsback and Zisserman(2008)]{nilsback2008automated}
Maria-Elena Nilsback and Andrew Zisserman.
\newblock {Automated flower classification over a large number of classes}.
\newblock In \emph{Indian Conference on Computer Vision, Graphics \& Image Processing}, pages 722--729. IEEE, 2008.

\bibitem[Papamarkou et~al.(2024)Papamarkou, Skoularidou, Palla, Aitchison, Arbel, Dunson, Filippone, Fortuin, Hennig, Hern{\'a}ndez-Lobato, et~al.]{papamarkou2024position}
Theodore Papamarkou, Maria Skoularidou, Konstantina Palla, Laurence Aitchison, Julyan Arbel, David Dunson, Maurizio Filippone, Vincent Fortuin, Philipp Hennig, Jos{\'e}~Miguel Hern{\'a}ndez-Lobato, et~al.
\newblock Position: Bayesian deep learning is needed in the age of large-scale ai.
\newblock \emph{arXiv preprint arXiv:2402.00809}, 2024.

\bibitem[Parkhi et~al.(2012)Parkhi, Vedaldi, Zisserman, and Jawahar]{parkhi2012cats}
Omkar~M Parkhi, Andrea Vedaldi, Andrew Zisserman, and CV Jawahar.
\newblock {Cats and dogs}.
\newblock In \emph{IEEE Conference on Computer Vision and Pattern Recognition}, pages 3498--3505. IEEE, 2012.

\bibitem[Qu et~al.(2025)Qu, Tao, Gong, Qu, Chen, Zhang, Wang, and Ding]{Qu2025-rx}
Zhen Qu, Xian Tao, Xinyi Gong, Shichen Qu, Qiyu Chen, Zhengtao Zhang, Xingang Wang, and Guiguang Ding.
\newblock Bayesian prompt flow learning for zero-shot anomaly detection.
\newblock \emph{arXiv [cs.CV]}, 2025.

\bibitem[Radford et~al.(2021)Radford, Kim, Hallacy, Ramesh, Goh, Agarwal, Sastry, Askell, Mishkin, Clark, et~al.]{radford2021learning}
Alec Radford, Jong~Wook Kim, Chris Hallacy, Aditya Ramesh, Gabriel Goh, Sandhini Agarwal, Girish Sastry, Amanda Askell, Pamela Mishkin, Jack Clark, et~al.
\newblock Learning transferable visual models from natural language supervision.
\newblock In \emph{International conference on machine learning}, pages 8748--8763. PmLR, 2021.

\bibitem[Recht et~al.(2019)Recht, Roelofs, Schmidt, and Shankar]{recht2019imagenet}
Benjamin Recht, Rebecca Roelofs, Ludwig Schmidt, and Vaishaal Shankar.
\newblock {Do ImageNet classifiers generalize to ImageNet?}
\newblock In \emph{International Conference on Machine Learning}, pages 5389--5400. PMLR, 2019.

\bibitem[Smith(1997)]{smith1997statistics}
Paul~J Smith.
\newblock Statistics: A bayesian perspective, 1997.

\bibitem[Soomro et~al.(2012)Soomro, Zamir, and Shah]{soomro2012ucf101}
Khurram Soomro, Amir~Roshan Zamir, and Mubarak Shah.
\newblock {UCF101: A dataset of 101 human actions classes from videos in the wild}.
\newblock \emph{arXiv preprint arXiv:1212.0402}, 2012.

\bibitem[Vaswani et~al.(2017)Vaswani, Shazeer, Parmar, Uszkoreit, Jones, Gomez, Kaiser, and Polosukhin]{vaswani2017attention}
Ashish Vaswani, Noam Shazeer, Niki Parmar, Jakob Uszkoreit, Llion Jones, Aidan~N Gomez, {\L}ukasz Kaiser, and Illia Polosukhin.
\newblock Attention is all you need.
\newblock \emph{Advances in neural information processing systems}, 30, 2017.

\bibitem[Villani et~al.(2008)]{villani2008optimal}
C{\'e}dric Villani et~al.
\newblock \emph{Optimal transport: old and new}.
\newblock Springer, 2008.

\bibitem[Wang et~al.(2019)Wang, Ge, Lipton, and Xing]{wang2019learning}
Haohan Wang, Songwei Ge, Zachary Lipton, and Eric~P Xing.
\newblock {Learning robust global representations by penalizing local predictive power}.
\newblock In \emph{Advances in Neural Information Processing Systems}, pages 10506--10518, 2019.

\bibitem[Welling and Teh(2011)]{welling2011bayesian}
Max Welling and Yee~W Teh.
\newblock Bayesian learning via stochastic gradient langevin dynamics.
\newblock In \emph{Proceedings of the 28th international conference on machine learning (ICML-11)}, pages 681--688, 2011.

\bibitem[Xiao et~al.(2010)Xiao, Hays, Ehinger, Oliva, and Torralba]{xiao2010sun}
Jianxiong Xiao, James Hays, Krista~A Ehinger, Aude Oliva, and Antonio Torralba.
\newblock {Sun database: Large-scale scene recognition from abbey to zoo}.
\newblock In \emph{IEEE Conference on Computer Vision and Pattern Recognition}, pages 3485--3492. IEEE, 2010.

\bibitem[Yang et~al.(2025)Yang, Zhang, Zhao, Luo, Zhong, Peng, and Fan]{Yang2025-fi}
Shijun Yang, Xiang Zhang, Wanqing Zhao, Hangzai Luo, Sheng Zhong, Jinye Peng, and Jianping Fan.
\newblock Multi-modal mutual-guidance conditional prompt learning for vision-language models.
\newblock \emph{arXiv [cs.CV]}, 2025.

\bibitem[Zhang et~al.(2019)Zhang, Li, Zhang, Chen, and Wilson]{zhang2019cyclical}
Ruqi Zhang, Chunyuan Li, Jianyi Zhang, Changyou Chen, and Andrew~Gordon Wilson.
\newblock Cyclical stochastic gradient mcmc for bayesian deep learning.
\newblock \emph{arXiv preprint arXiv:1902.03932}, 2019.

\bibitem[Zhang et~al.(2021)Zhang, Fang, Zhang, Gao, Li, Dai, Qiao, and Li]{tipadapter}
Renrui Zhang, Rongyao Fang, Wei Zhang, Peng Gao, Kunchang Li, Jifeng Dai, Yu Qiao, and Hongsheng Li.
\newblock Tip-adapter: Training-free clip-adapter for better vision-language modeling.
\newblock \emph{arXiv preprint arXiv:2111.03930}, 2021.

\bibitem[Zhou et~al.(2022{\natexlab{a}})Zhou, Yang, Loy, and Liu]{zhou2022conditional}
Kaiyang Zhou, Jingkang Yang, Chen~Change Loy, and Ziwei Liu.
\newblock Conditional prompt learning for vision-language models.
\newblock In \emph{Proceedings of the IEEE/CVF conference on computer vision and pattern recognition}, pages 16816--16825, 2022{\natexlab{a}}.

\bibitem[Zhou et~al.(2022{\natexlab{b}})Zhou, Yang, Loy, and Liu]{zhou2022learning}
Kaiyang Zhou, Jingkang Yang, Chen~Change Loy, and Ziwei Liu.
\newblock Learning to prompt for vision-language models.
\newblock \emph{International Journal of Computer Vision}, 130\penalty0 (9):\penalty0 2337--2348, 2022{\natexlab{b}}.

\bibitem[Zhu et~al.(2025)Zhu, Niu, Han, Wu, and Zhang]{Zhu2025-jq}
Beier Zhu, Yulei Niu, Yucheng Han, Yue Wu, and Hanwang Zhang.
\newblock Prompt-aligned gradient for prompt tuning.
\newblock \emph{arXiv [cs.CV]}, 2025.

\end{thebibliography}
}

\appendix
\clearpage
\setcounter{page}{1}
\maketitlesupplementary


\section{Additional Details and Experiments}\label{appx:additional-experiments}

\noindent\textbf{Implementation Details.} We utilize the ViT-B/16 variant of CLIP~\cite{radford2021learning} as our vision-language backbone. All experiments are executed on NVIDIA L4 and L40S GPUs. We report average accuracy on base classes, novel classes, and their harmonic mean, computed over three independent runs with different random seeds, to ensure statistical reliability, following the experimental protocol for prompt learning methods~\cite{zhou2022conditional, khattak2023maple}. Since the MMRL method uses the AdamW optimizer, for our MMRL + ReBaPL method we perform a AdamW burn-in during the initial iterations of each cycle to reach the neighborhood of the local mode, and then switch to SGHMC in the final iterations of the cycle.

\subsection{Toy Example}\label{appx:toy}

We start with a toy example to illustrate our method. We consider a simple Gaussian mixture potential $U(\theta) = \frac{1}{2}(\mathcal{N}(\cdot|\mu_{1},\Sigma_{1}) + \mathcal{N}(\cdot|\mu_{2},\Sigma_{2}))$ and $V(\theta, \theta')$ as in Equation~\ref{eq:potential}, with $d(\theta, \theta') = \lVert \theta - \theta' \rVert_{2}$ for simplicity. The underlying potential $U$ has two modes ($\mu_{1}$ and $\mu_{2}$), and we want to explore both of them through MCMC (SGLD in this case, for simplicity).

\begin{figure}[ht]
    \centering
    \begin{subfigure}{0.333\linewidth}
        \centering
        \includegraphics[width=\linewidth]{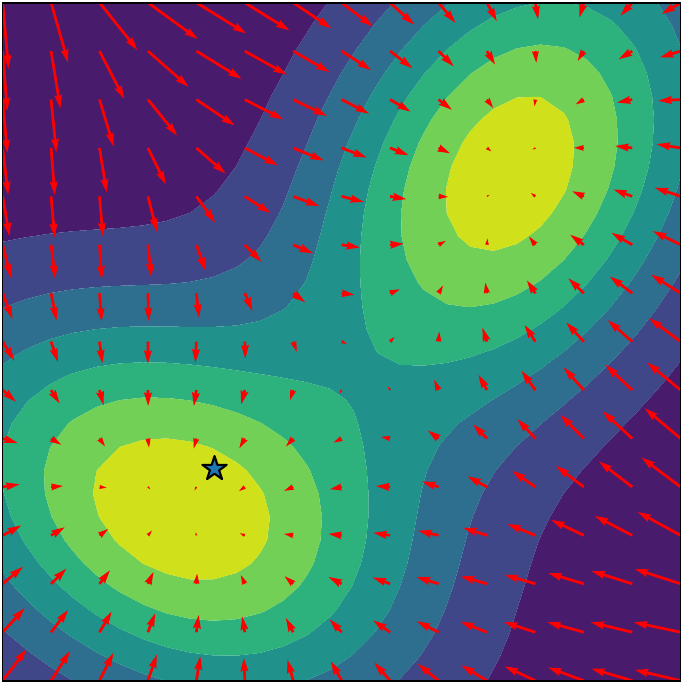}
        \caption{Potential $U(\theta)$}
    \end{subfigure}\hfill
    \begin{subfigure}{0.333\linewidth}
        \centering
        \includegraphics[width=\linewidth]{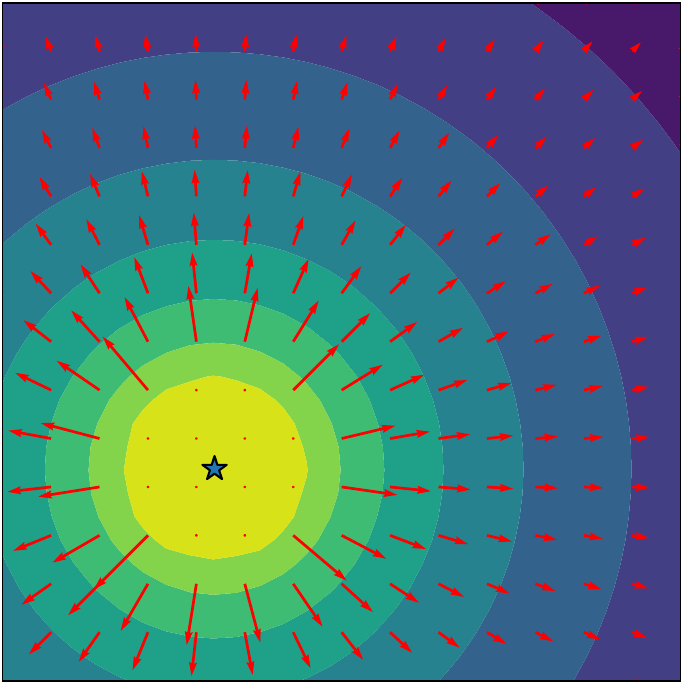}
        \caption{Potential $V(\theta, \theta')$}
    \end{subfigure}\hfill
    \begin{subfigure}{0.333\linewidth}
        \centering
        \includegraphics[width=\linewidth]{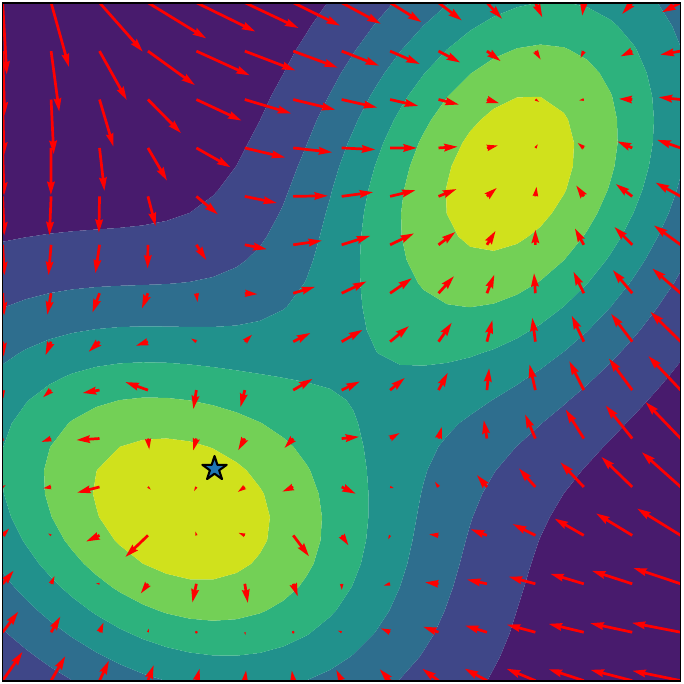}
        \caption{Field w/ repulsion}
    \end{subfigure}\hfill
    \caption{Conceptual illustration of the benefits of repulsive MCMC on a mixture of Gaussian distributions. In (a), we show the potential $U(\theta)$ alongside the vector field $\nabla U(\theta)$. The blue point, $\theta_{k,T}^{(c-1)}$ represents the minimum of $\theta \mapsto U(\theta)$ obtained through \gls{sgld}. In (b), we show the repulsion potential $V(\theta,\theta_{k,T}^{(c-1)})$ alongside the vector field $F(\theta,\theta_{k,T}^{(c-1)}) = \nabla_{\theta}V(\theta,\theta_{k,T}^{(c-1)})$. As we show in (c) the repulsion vector field changes the initial vector field, pushing samples away from the sample of the previous cycle.}
    \label{fig:toy1}
\end{figure}

\begin{figure}[ht]
    \centering
    \begin{subfigure}{0.499\linewidth}
        \centering
        \includegraphics[width=\linewidth]{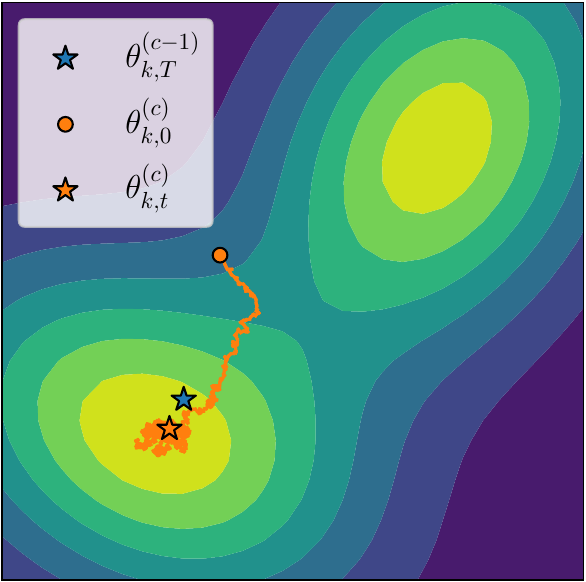}
        \caption{SGLD}
    \end{subfigure}\hfill
    \begin{subfigure}{0.499\linewidth}
        \centering
        \includegraphics[width=\linewidth]{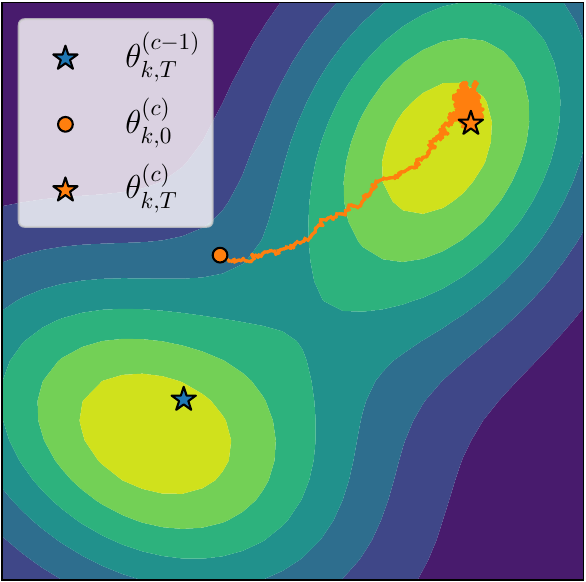}
        \caption{SGLD + Repulsion}
    \end{subfigure}\hfill
    \caption{As we show in (a) and (b), we encourage mode exploration by driving samples from the current cycle, $\theta_{k,t}^{(c)}$, away from those of the previous cycle, $\theta_{k,T}^{(c-1)}$.}
    \label{fig:toy2}
\end{figure}

We start by running standard \gls{sgld} to obtain a sample in the posterior, denoted by the blue star in Figures~\ref{fig:toy1} and~\ref{fig:toy2}. As shown in Figure~\ref{fig:toy1} (a), this point is located in a high density region of $U(\theta)$ . For the next cycle, we add repulsion for exploring the second mode in the landscape of $U(\theta)$. The repulsion potential, $V$, is shown in Figure~\ref{fig:toy1} (b). Note how the vector field, i.e., $F(\theta,\theta_{k,T}^{(c-1)})$, point outwards from $\theta_{k,T}^{(c-1)}$. We show the posterior alongside the final vector field $\nabla_{\theta} \biggl(U(\theta) + \xi V(\theta,\theta_{k,T}^{(c-1)})\biggr)$ in Figure~\ref{fig:toy1} (c). Overall, repulsion modifies the vector field, driving samples away from $\theta_{k,T}^{(c-1)}$, thus enhancing mode exploration.

We then run \gls{sgld} and \gls{sgld} with repulsion on this toy example. The results are shown in Figure~\ref{fig:toy2} (a) and (b) from an initialization $\theta_{k,0}^{(c)}$. Due the initialization, \gls{sgld} converges to the same mode as $\theta_{k,T}^{(c-1)}$. In comparison, by adding the repulsion term, $\theta_{k,0}^{(c)}$ converges to the second mode in the posterior.

\begin{remark}{(Intuition on repulsion strength $\xi$)}
    The repulsion strength $\xi$ strikes a balance between the force driving samples to the modes of the log-likelihood, i.e., $\nabla_{\theta}U(\theta)$, and the repulsive force driving samples away from previous cycles' sample, i.e., $\nabla_{\theta}V(\theta,\theta_{k,T}^{(c-1)})$. Intuitively, if $\xi \rightarrow +\infty$, the repulsive strength overpowers the force towards the modes of the posterior. Therefore, tuning $\xi$ is important to strike a balance between standard SGMCMC and mode exploration. We show in Figure~\ref{fig:big_v} an illustration of this phenomenon.

    \begin{figure}[ht]
        \centering
        \includegraphics[width=0.49\linewidth]{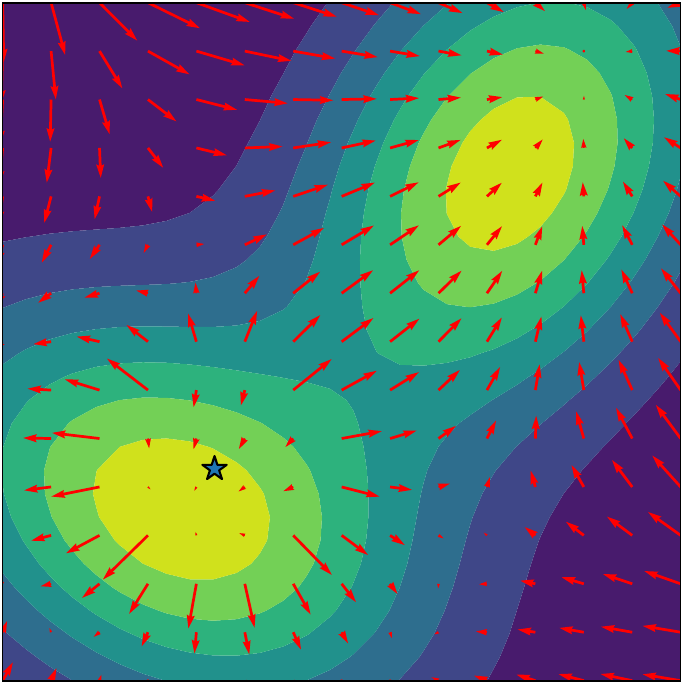}\hfill
        \includegraphics[width=0.49\linewidth]{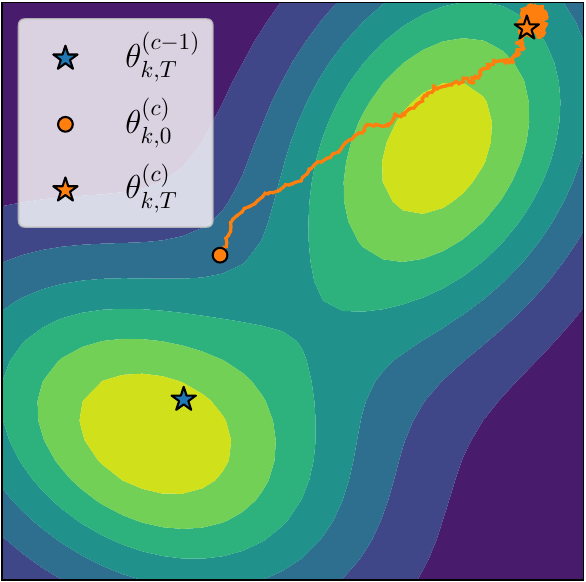}
        \caption{SGMCMC with very high repulsion strength $\xi$. In this case the repulsive force dominates the sampling trajectory.}
        \label{fig:big_v}
    \end{figure}
\end{remark}

\subsection{Prompt Diversity}

An important aspect of MCMC is yielding diverse samples from the posterior distribution. This can be analyzed through the lens of mode collapse, i.e., whether the samples $\{\theta_{k,T}^{(c)}\}_{k=1}^{K}$ are close with respect to some defined metric.

To assess whether our method successfully mitigates mode collapse, we analyzed the similarity in feature distributions between the sampled models. We computed the Wasserstein distance between the representations for three independent checkpoints. This metric quantifies how distinct the internal feature distributions are between models, with a higher distance indicating greater diversity between the feature distributions.

    \begin{figure}[t]
    \centering
    \begin{subfigure}{0.499\linewidth}
        \centering
        \includegraphics[width=\linewidth]{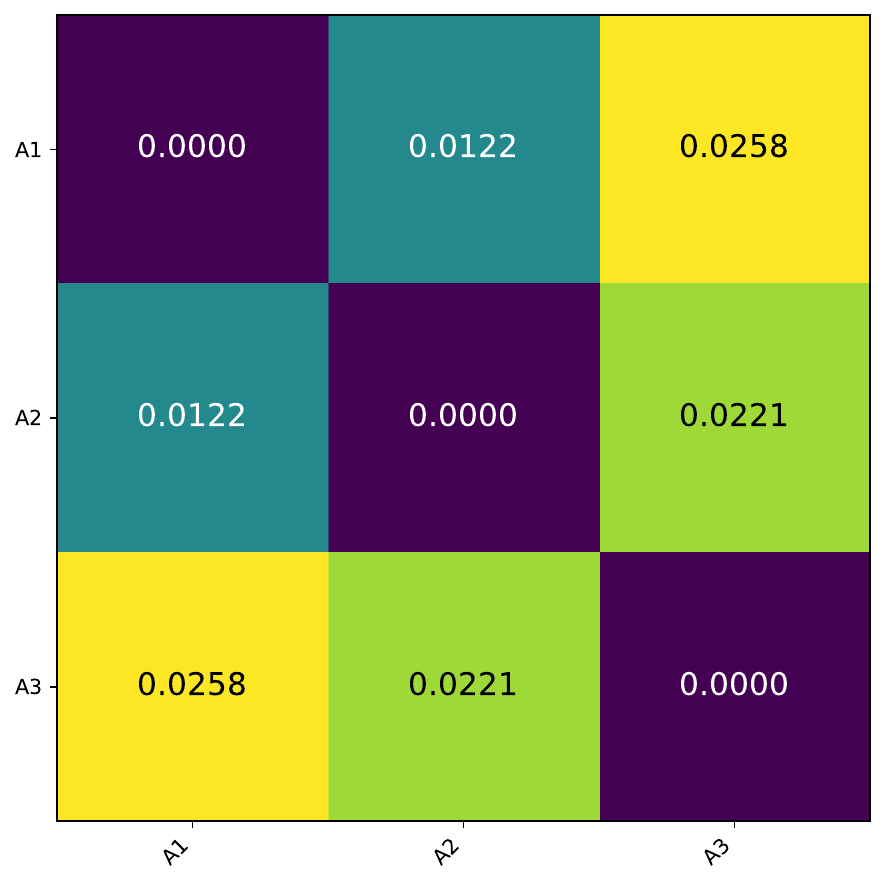}
        \caption{Training A with repulsion}
    \end{subfigure}\hfill
    \begin{subfigure}{0.499\linewidth}
        \centering
        \includegraphics[width=\linewidth]{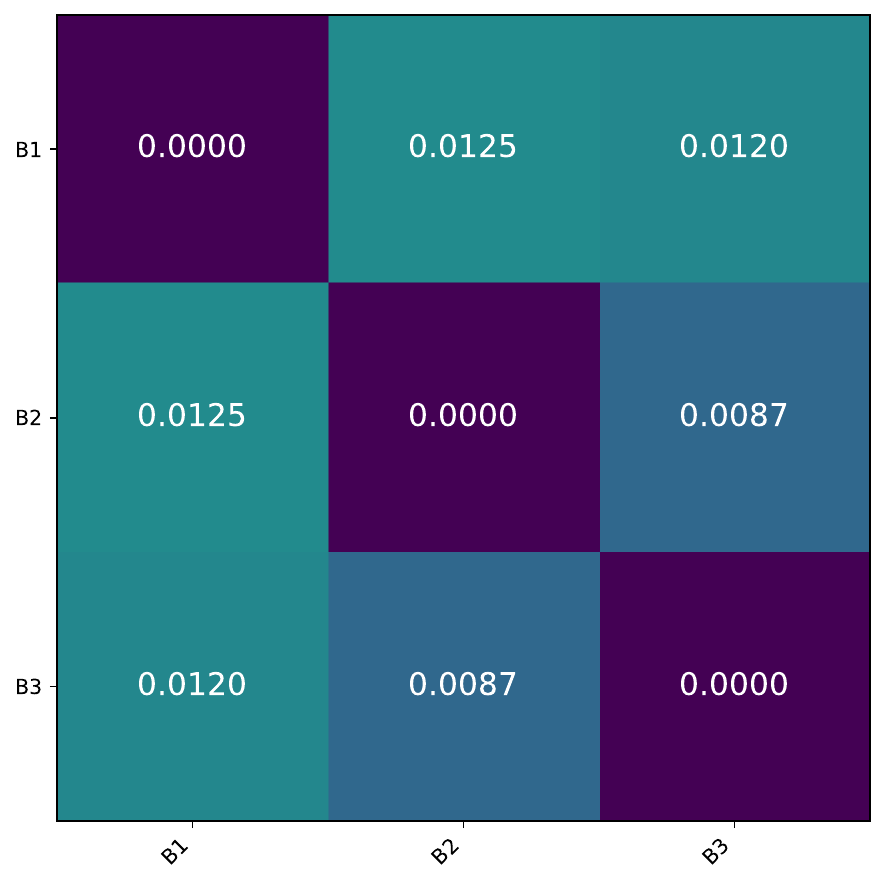}
        \caption{Training B without repulsion}
    \end{subfigure}\hfill
    \caption{Wasserstein distance matrix after training on Eurosat, with and without repulsion. Average Wasserstein distance with repulsion is greater.}
    \label{fig:wasserstein_heatmap}
    \end{figure}

To quantify the diversity of the learned features, we compute the pairwise Wasserstein distances between the latent representations of models from different training cycles. Using a test batch of 100 images, we construct a matrix $D$ with entries $D_{i,j} = W_2(U_{\theta_i}, U_{\theta_j})$. Here, $U_{\theta_i}$ is defined as the empirical distribution of the latent features associated with cycle $i$.

As illustrated in Figure~\ref{fig:wasserstein_heatmap}, the resulting distance matrices reveal the extent of functional separation between samples. We observe that for Training A, which employs repulsion, the Wasserstein distances between the representation distributions across different cycles are, on average, greater than those observed in Training B.

\subsection{Ablations}

In this section we ablate the hyperparameters of our method, including the batch size $n$ used to calculate the repulsion force, the repulsion strength $\xi$, the number of cycles $C$, and the number of posterior samples $K$.

\begin{figure}[ht]
    \centering
    \includegraphics[width=\linewidth]{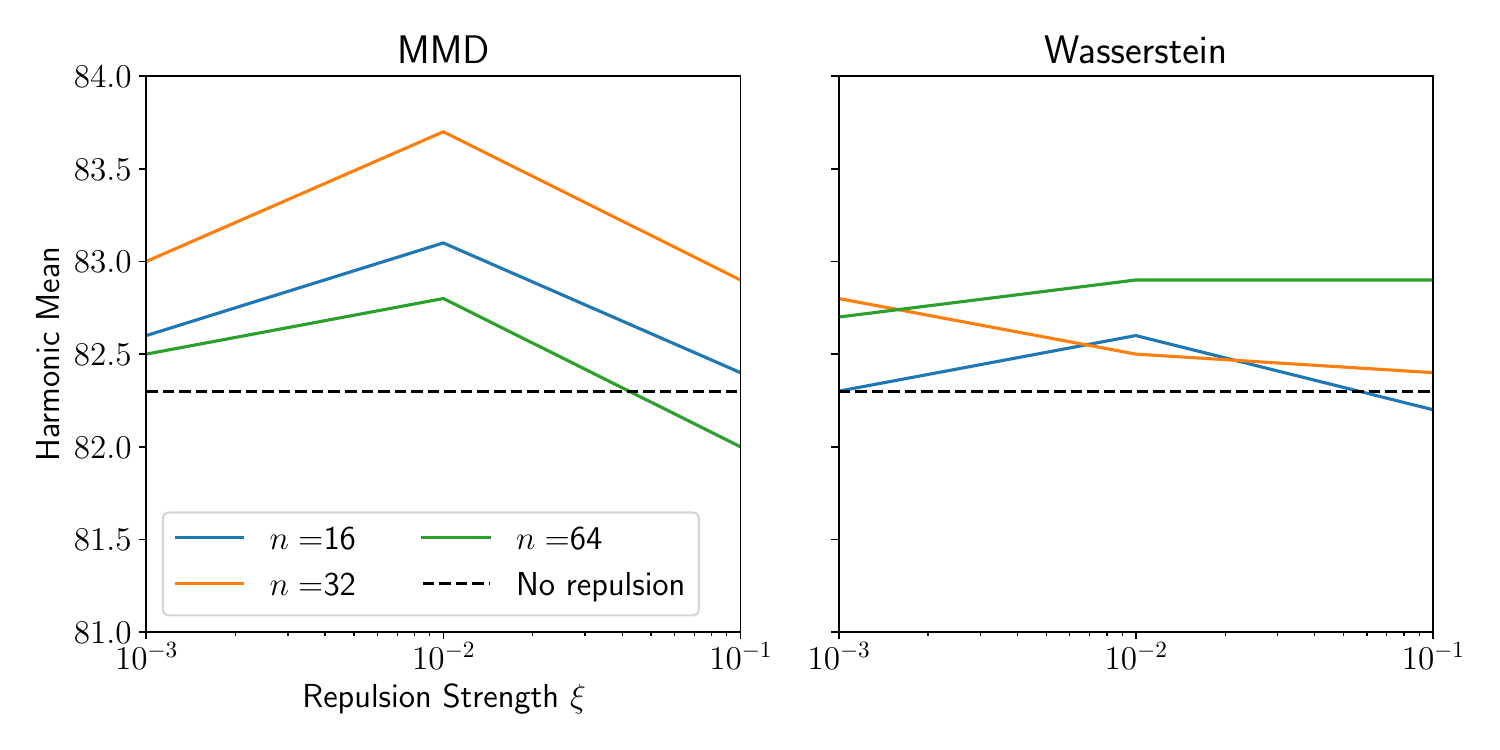}
    \caption{Harmonic mean of base and novel accuracies on the UCF dataset, as a function of repulsion strength $\xi$ for various batch sizes.}
    \label{fig:ablation_xi_hm}
\end{figure}
\noindent\textbf{Batch size $n$ and repulsion strength $\xi$.} We use $n \in \{16, 32, 64\}$, $\xi \in \{10^{-3}, 10^{-2},  10^{-1}\}$, and $\text{distance} \in \{\text{MMD},\text{Wasserstein}\}$, with a total of $18$ combinations. We show our results in Figure~\ref{fig:ablation_xi_hm} for the UCF dataset. Overall, this figure shows the inherent trade-off associated with repulsion strength. On the one hand, a small repulsion strength is not enough to encourage exploration, resulting in a smaller improvement in performance. On the other hand, large repulsion leads to too much exploration, as we discussed in our toy example (Appendix~\ref{appx:toy}).


\begin{figure}[t]
    \centering
    \includegraphics[width=0.6\linewidth]{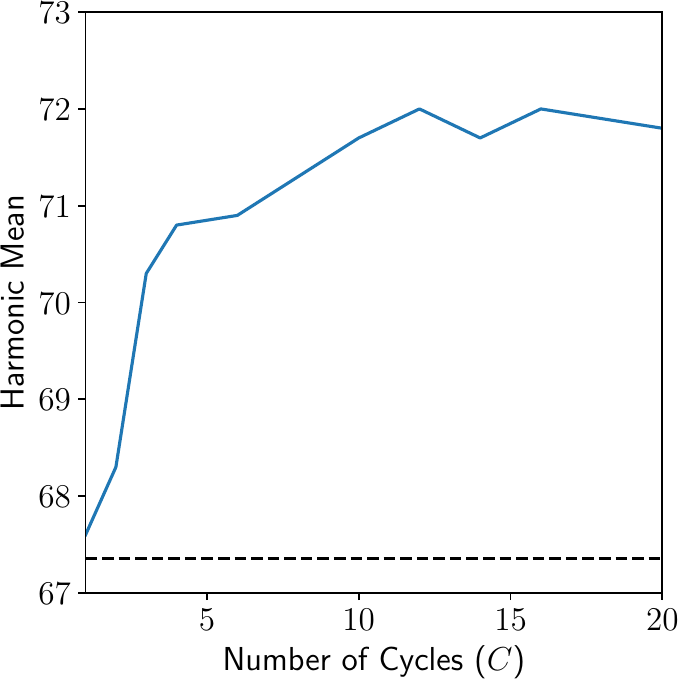}
    \caption{Ablation on the number of cycles in the rcSGHMC algorithm for the DTD dataset.}
    \label{fig:ablation_n_cycles}
\end{figure}
\noindent\textbf{Number of cycles $C$.} For $C \in \{1, 2, 3, 4, 6, \cdots, 20\}$ in our algorithm. We show our results in Figure~\ref{fig:ablation_n_cycles}. First, repulsive cSGMCMC improves over the MaPLe baseline for all number of cycles. Second, running Algorithm~\ref{alg:csghmc} for more cycles generally leads to better performance. This is somewhat intuitive, since we are able to acquire more samples from additional modes in the posterior distribution. These findings show that, through repulsion, our method is able to better explore the posterior landscape.

\newpage



\begin{wraptable}{r}{0.5\linewidth}
\centering
\resizebox{\linewidth}{!}{
\begin{tabular}{cccc}
    \toprule
    \# Postrior Samples $K$ & Base & Novel & HM \\
    \midrule
    1 & 97.43 & 73.93 & 84.07\\
    2 & 97.30 & 74.10 & 84.12\\
    3 & 97.30 & 74.13 & 84.14\\
    4 & 97.37 & 74.17 & 84.19\\
    5 & 97.33 & 74.23 & 84.22\\
    \bottomrule
\end{tabular}
}
\caption{MaPLe + ReBaPL classification accuracy (\%) vs. $K$, avg. over 3 seeds on Flowers102.}
\label{tab:spc-ablation}
\vspace{-1em}
\end{wraptable}
\noindent\textbf{Number of posterior samples.} First, we ensemble over $\{ \{ \theta_{k,T}^{(c)} \}_{c=1}^{C} \}_{k=1}^{K}$, i.e., $K$ samples from $C$ cycles. Our results in Table~\ref{tab:main_results} use $K=1,C=3$ (3 total samples). Figure~\ref{fig:ablation_n_cycles} shows the baseline $C = K = 1$ (single sample), where performance is $\approx 67.5\%$ HM, marginally above the no repulsion baseline (dashed line, $\approx 67.4\%$). Increasing $C$ yields clear gains: $\approx 70.5$ at $C = 3$ up to $\approx 72\%$ at $C = 10$, before saturating. This ablation demonstrates that multiple posterior samples ($K=1$ from each cycle) are necessary to achieve the reported performance. For completeness, Table~\ref{tab:spc-ablation} explores performance with varying $K$ on Flower102, showing that performance is stable and gains come primarily from $C$. We use $K = 1$ in the main paper for efficiency.

\subsection{Overhead of computing probability metrics}\label{appx:overhead}

\begin{wraptable}{r}{0.48\linewidth}
\vspace{-1em}
\centering
\resizebox{\linewidth}{!}{
\begin{tabular}{ccc}
\toprule
Batch size & MMD & Wasserstein \\
\midrule
2  & 0.381 & 0.496 \\
4  & 0.382 & 0.523 \\
8  & 0.396 & 0.558 \\
16 & 0.538 & 0.525 \\
32 & 0.890 & 0.657 \\
\bottomrule
\end{tabular}
}
\caption{Running time (in milliseconds) of probability metrics for different batch sizes.}
\label{tab:distance-overhead}
\vspace{-0.3cm}
\end{wraptable}
The use of probability metrics such as the MMD or the Wasserstein distance does not introduce much overhead, since we compute them on mini-batches of cached representations from the MCMC samples. For reference, computing \gls{mmd} has computational complexity $\mathcal{O}(n^{2})$, where $n$ is the number of samples in the support of the distribution. The Wasserstein distance has $\mathcal{O}(n^{3})$ complexity for exact solvers based on linear programming. We refer readers to~\cite{montesuma2024recent} for more details on these complexities. Table~\ref{tab:distance-overhead} provides a comparison of the MMD and Wasserstein distance running time vs. \# of representation samples. In comparison, the entire computational overhead involved in repulsion is $\approx$ 700 ms, related to the forward and backward passes. Hence, the overhead is negligible.

\subsection{Efficiency Analysis and Inference Parallelism}\label{appx:efficiency}

\begin{wraptable}{r}{0.5\linewidth}
\vspace{-1em}
\centering
\resizebox{\linewidth}{!}{
\begin{tabular}{ccc}
\toprule
Cycles ($C$) & MaPLe & MaPLe + ReBaPL \\
\midrule
1 & 242.33 & 242.39 \\
2 & --                & 259.35 \\
3 & --                & 276.50 \\
4 & --                & 298.91 \\
\bottomrule
\end{tabular}
}
\caption{Avg. inference latency (in miliseconds) over 20 runs.}
\label{tab:inference-latency}
\vspace{-1em}
\end{wraptable}
ReBaPL trains for $E \times C$ total epochs ($E=5,C=3$). Training MaPLe for 15 epochs yields similar times (10m40s vs. 10m54s), as repulsion uses cached features ($\approx$ 700ms/iter). Critically, MaPLe at 15 epochs \textbf{overfits}: Base/Novel/HM of 97.23/70.90/82.00 vs. 96.33/73.33/83.27 at 5 epochs. Thus, the gain stems from ReBaPL itself, not more compute. Furthermore, ReBaPL inference parallelizes with minimal overhead; see Table~\ref{tab:inference-latency} for varying $C$ when running on 4 L4 GPUs.

\section {Additional Implementation Details}
We describe in Table~\ref{tab:hyperparameters} the set of hyperparameters used to run our experiments in Table~\ref{tab:main_results}.
\begin{table}[ht]
    \footnotesize
    \centering
    \caption{Hyperparameters used in our ReBaPL experiments for both MMRL and MaPLe.}
    \begin{tabular}{lcc}
        \toprule
         Hyperparameter & MaPLe  & MMRL\\
         \midrule
         Learning rate $\alpha$ & 0.002 & 0.001\\
         Batch size $b$ & 1 & 4\\
         Epochs per cycle $T$ & 5 & 5\\
         \# Cycles $C$ & 3 & 3\\
         Samples per cycle $K$ & 1 & 1\\
         Repulsion strength $\xi$ & 0.001 & $10^{-4}$\\
         \textbf{Repulsion} batch size $n$ & 32 & 64\\
         Distance & MMD & Wasserstein\\
         Algorithm & \multicolumn{2}{c}{rcSGHMC}  \\
         Weight decay & \multicolumn{2}{c}{5e-4} \\
         Random seed & \multicolumn{2}{c}{[1, 2, 3]} \\
         \bottomrule
    \end{tabular}
    \label{tab:hyperparameters}
\end{table}

\section{Additional Background}\label{appx:background}

\subsection{Probability Metrics}

We give further details on the probability metrics introduced in Section~\ref{sec:prob-metrics}, especially Equations~\ref{eq:mmd} and~\ref{eq:w2}. Here, we cover the essential theory behind the \gls{mmd} and the Wasserstein distance. For further details, we refer readers to~\cite{gretton2012kernel} and~\cite{montesuma2024recent}.

\noindent\textbf{Maximum Mean Discrepancy.} Let $\mathcal{F}$ be a family of test functions over $\mathcal{U}$. The \gls{mmd} is defined by,
\begin{align*}
    \text{MMD}(p, q) = \underset{f \in \mathcal{F}}{\text{sup}} \mathbb{E}_{x\sim p}[f(x)] - \mathbb{E}_{y \sim q}[f(y)].
\end{align*}
When $\mathcal{F}$ is a reproducing kernel Hilbert space (see~\cite[Section 2.2]{gretton2012kernel}) with kernel $\kappa:\mathcal{U}\times\mathcal{U}\rightarrow\mathbb{R}$, the squared \gls{mmd} is expressed in terms of the mean embeddings of $p$ and $q$,
\begin{align*}
    \text{MMD}(p, q)^{2} = \lVert \mu_{p}-\mu_{q} \rVert_{2}^{2}.
\end{align*}
Here, the mean embedding is the element $\mu_{p} \in \mathcal{F}$ such that $\langle f, \mu_{p} \rangle_{\mathcal{F}} = \mathbb{E}_{x \sim p}[f]$, $\forall f \in \mathcal{F}$. By~\cite[Lemma 6]{gretton2012kernel}, the squared \gls{mmd} admits the form,
\begin{align}
    \text{MMD}(p, q)^{2} = \underset{x \sim p,x' \sim p}{\mathbb{E}}[\kappa(x,x')] +& \underset{y \sim q,y' \sim q}{\mathbb{E}}[\kappa(y,y')]\nonumber\\ -& 2\underset{x \sim p,y \sim q}{\mathbb{E}}[\kappa(x,y)],\label{eq:true_mmd}
\end{align}
which then admits the \emph{unbiased} estimator in Equation~\ref{eq:mmd}. Due to the double summations in Equation~\ref{eq:mmd}, and with the assumption that $n \approx m$, the computational complexity of computing the empirical \gls{mmd} is $\mathcal{O}(n^{2})$.

\noindent\textbf{Wasserstein distance.} This distance is rooted in the theory of \gls{ot}~\cite{villani2008optimal,montesuma2024recent}, and gives the \emph{least amount of effort or energy} required to move the probability mass from $p$ to $q$. In other words, consider the set $\Gamma(p, q)$ so that $\gamma \in \Gamma(p, q)$ satisfies,
\begin{align*}
    \int_{\mathcal{U}}\gamma(x,y)dy = p(x)\quad\int_{\mathcal{U}}\gamma(x,y)dx = q(y).
\end{align*}
The elements of $\gamma$ are called \emph{transport plans}, and can be conceptualized as joint distributions with marginals $p$ and $q$. Note that $\Gamma(p, q)$ also impose mass preservation constraints on the transportation plans. With these concepts,
\begin{align}
    \gamma^{\star} = \underset{\gamma \in \Gamma(p, q)}{\text{arginf}}\int_{\mathcal{U}}\int_{\mathcal{U}}c(x,y)\gamma(x,y)dxdy,\label{eq:kantorovich}
\end{align}
where $c:\mathcal{U}\times\mathcal{U}\rightarrow\mathbb{R}$ is called the \emph{ground-cost}, i.e., a function that measures the effort of moving $x$ to $y$.

The problem in Equation~\ref{eq:kantorovich} can induce a metric on $\mathcal{P}(\mathcal{U})$ under certain conditions. Especially, let $(\mathcal{U}, d_{\mathcal{U}})$ be a metric space, and $c(x, y) = d(x, y)^{\alpha}$, $\alpha \in [1, +\infty)$. The \gls{ot} cost under these conditions defines the $\alpha-$Wasserstein distance,
\begin{align}
    W_{\alpha}(p, q)^{\alpha} = \underset{\gamma \in \Gamma}{\text{inf}}\int_{\mathcal{U}}\int_{\mathcal{U}}d(x,y)^{\alpha}\gamma(x, y)dxdy.\label{eq:true_w2}
\end{align}
This is an infinite dimensional program on the variable $\gamma$. Given samples $x_i^{(p)} \sim p$ and $y_{j}^{(q)} \sim q$, the Wasserstein distance admits an empirical estimator given by Equation~\ref{eq:w2}. In this case, it is calculated through a finite linear program with $n \times m$ variables. In that sense, its computational complexity is $\mathcal{O}(n^{3})$.

\begin{remark}{(Entropic Regularization)}
    An alternative to Equation~\ref{eq:w2} is to consider the Sinkhorn divergence~\cite{cuturi2013sinkhorn,genevay2018learning}, which is a regularized version of the \gls{ot} problem, which has $\mathcal{O}(n^{2})$ complexity \emph{per iteration}. However, in small sample scenarios (e.g., at the level of a mini-batch), Sinkhorn divergence usually needs many iterations to accurately estimate the transportation plan, which leads to a running time that is comparable or worse than exact \gls{ot}. For that reason, in our experiments, we use the standard Wasserstein distance.
\end{remark}

\begin{remark}{(Mini-batch estimation)}
    In the main paper (c.f. Equations~\ref{eq:mmd} and~\ref{eq:w2}), we presented the empirical estimators of the continuous \gls{mmd} and Wasserstein distance (c.f. Equations~\ref{eq:true_mmd} and~\ref{eq:true_w2}). These estimators assume $n$ and $m$ i.i.d. samples from distributions $p$ and $q$, respectively. Now, note that our main interest is computing $d_{\mathcal{P}(\mathcal{U})}(U_{\theta}, U_{\theta'})$, where $U_{\theta} = \{u_{\theta,i}\}_{i=1}^{n}$ and $U_{\theta'} = \{u_{\theta', i}\}_{i=1}^{n}$. In other words, we understand $U_{\theta}$ and $U_{\theta'}$ as the i.i.d. samples from the distribution of representations produced by networks $\theta$ and $\theta'$, respectively.
    
    For the sake of completeness, the \gls{mmd} reads as,
    \begin{align*}
        \text{MMD}(U_{\theta}, U_{\theta'})^{2} = \dfrac{1}{n^{2}}\biggr( &\sum_{i,j}\kappa(u_{\theta,i},u_{\theta,j}) +\\ &\sum_{i,j}\kappa(u_{\theta',i},u_{\theta',j}) -\\ &2\sum_{i,j}\kappa(u_{\theta,i},u_{\theta',j})\biggr),
    \end{align*}
    and the Wasserstein distance,
    \begin{align*}
        W_{2}(U_{\theta}, U_{\theta'})^{2} = \sum_{i=1}^{n}\sum_{j=1}^{n}\gamma_{ij}^{\star}\lVert u_{\theta,i} - u_{\theta',j} \rVert_{2}^{2},
    \end{align*}
    where $\gamma^{\star}$ is obtained through linear programming.
\end{remark}



\end{document}